\documentclass[
11pt,
oneside,
english, 
singlespacing, 
headsepline, 
]{MastersDoctoralThesis} 

\usepackage[utf8]{inputenc} 
\usepackage[T1]{fontenc} 

\usepackage{mathpazo} 
\usepackage{listings}

\usepackage[backend=bibtex,style=authoryear,natbib=true]{biblatex} 

\usepackage[autostyle=true]{csquotes} 

\thesistitle{Deep Learning with Tabular Data:\\A Self-supervised Approach} 
\supervisor{Dr. Lingqiao \textsc{Liu}} 
\examiner{} 
\degree{Master's of Computer Science} 
\author{Tirth K \textsc{Vyas}} 
\addresses{} 

\subject{Computer Sciences} 
\keywords{} 
\university{\href{https://www.adelaide.edu.au/}{The University Of Adelaide}} 
\department{School of Computer and Mathematical Sciences} 
\faculty{\href{https://set.adelaide.edu.au/computer-and-mathematical-sciences/disciplines/computer-science}{Computer Science}} 

\AtBeginDocument{
\hypersetup{pdftitle=\ttitle} 
\hypersetup{pdfauthor=\authorname} 
\hypersetup{pdfkeywords=\keywordnames} 
}

\begin{document}

\frontmatter 

\pagestyle{plain} 


\begin{titlepage}
\begin{center}

\vspace*{.06\textheight}
{\scshape\LARGE \univname\par}\vspace{1.5cm} 
\textsc{\Large Master's Thesis}\\[0.5cm] 

\HRule \\[0.4cm] 
{\huge \bfseries \ttitle\par}\vspace{0.4cm} 
\HRule \\[1.5cm] 
 
\begin{minipage}[t]{0.4\textwidth}
\begin{flushleft} \large
\emph{Author:}\\
\href{https://tirthkvyas.github.io/}{\authorname}\\ 
\emph{ID: a1842834}
\end{flushleft}
\end{minipage}
\begin{minipage}[t]{0.4\textwidth}
\begin{flushright} \large
\emph{Supervisor:} \\
\href{https://lingqiao-adelaide.github.io/lingqiaoliu.github.io//}{\supname} 
\end{flushright}
\end{minipage}\\[3cm]
 
\vfill

\large \textit{A thesis submitted in fulfillment of the requirements for the course Research Project Part A and B for the degree of \degreename.}\\[0.3cm] 
 
\vfill

{\large \today}\\[4cm] 

\vfill
\end{center}
\end{titlepage}


\begin{declaration}
\addchaptertocentry{\authorshipname} 
\noindent I, \authorname, declare that this thesis titled, \enquote{\ttitle} and the work presented in it are my own. I confirm that:

\begin{itemize} 
\item This work was done wholly or mainly while in candidature for a research degree at this University.
\item Where any part of this thesis has previously been submitted for a degree or any other qualification at this University or any other institution, this has been clearly stated.
\item Where I have consulted the published work of others, this is always clearly attributed.
\item Where I have quoted from the work of others, the source is always given. With the exception of such quotations, this thesis is entirely my own work.
\item I have acknowledged all main sources of help.
\item Where the thesis is based on work done by myself jointly with others, I have made clear exactly what was done by others and what I have contributed myself.\\
\end{itemize}
 
\noindent Signed: Tirth Kiranbhai Vyas\\
\rule[0.5em]{25em}{0.5pt} 
 
\noindent Date: \today\\
\rule[0.5em]{25em}{0.5pt} 
\end{declaration}

\cleardoublepage


\vspace*{0.2\textheight}

\noindent\enquote{\itshape It’s not about how smart or talented you are, it’s just about how longer you can stick with your problems.}\bigbreak

\noindent\enquote{\itshape I have no special talents. I am only passionately curious.}\bigbreak

\hfill Albert Einstein


\begin{abstract}
\addchaptertocentry{\abstractname} 
In the real-world application disciplines like Computer Vision, NLP (Natural Language Processing), Machine Translation, Speech Recognition, Audio Identification, and Bioinformatics, deep learning has been demonstrating successful outcomes. And certainly, that’s because of the advancements in the domain of deep neural network architectures. However, deep-learning models haven't yet excelled in the domain of tabular datasets. Various machine-learning models are been providing state-of-the-art results, which are majorly based on regression or classification trees methodology as base learners. Therefore, there is a need to explore how deep learning can be applied on tabular data to improve accuracy and performance.\hfill\\\\
Hence in research, we have described a novel approach for training tabular data using the TabTransformer model with self-supervised learning. Traditional machine learning models for tabular data, such as GBDT are being widely used though our paper examines the effectiveness of the TabTransformer which is a Transformer-based model optimized specifically for tabular data. The TabTransformer captures intricate relationships and dependencies among features in tabular data by leveraging the self-attention mechanism of Transformers. I have used a self-supervised learning approach in this study, where the TabTransformer learns from unlabeled data by creating surrogate supervised tasks, eliminating the need for the labelled data. The aim is to find the most effective TabTransformer model representation of categorical and numerical features. To address the challenges faced during the construction of various input settings into the Transformers. Furthermore, a comparative analysis is also been conducted to examine performance of the TabTransformer model against baseline models such as MLP and supervised TabTransformer. \hfill\\\\
The comparison includes traditional supervised learning models and self-supervised learning approaches. Through this analysis, the effectiveness of self-supervised TabTransformer model is demonstrated, showcasing its potential against the traditional machine learning models and competing with the latest advancements in self-supervised learning approaches. Overall, my experimental study advances training tabular data by introducing the TabTransformer model and demonstrating its potential with a self-supervised learning approach. The findings shed light on the best way to represent categorical and numerical features, emphasizing the TabTransformer's performance when compared to established machine learning models and other self-supervised learning methods.\end{abstract}


\begin{acknowledgements}
\addchaptertocentry{\acknowledgementname} 
Special thanks to Dr. Lingqiao Liu and Qiaoyang Luo for their unwavering support, patience and expert guidance during the research which greatly helped in shaping the research's vision and making it possible.
\end{acknowledgements}


\tableofcontents 

\listoffigures 


\begin{abbreviations}{ll} 

\textbf{ML} & \text{Machine Learning}\\
\textbf{NLP} & \text{Natural Language Processing}\\ 
\textbf{DL} & \text{Deep Learning}\\
\textbf{SSL} & \text{Self-supervised Learning}\\
\textbf{SL} & \text{Supervised Learning}\\
\textbf{TT} & \text{TabTransformer}\\
\textbf{B-TT} & \text{Binned TabTransformer}\\
\textbf{VM-TT} & \text{Vanilla-MLP-TabTransformer}\\
\textbf{V-TT} & \text{Vanilla-TabTransformer}\\
\textbf{MLP} & \text{Multi-layer Perception}\\
\textbf{MLP-TT} & \text{MLP-based TabTransformer}\\
\textbf{GBDT} & \text{Gradient Boosted Decision Trees}\\
\textbf{MSE} & \text{Mean Squared Error}\\
\textbf{MAE} & \text{Mean Absolute Error}\\
\textbf{AI} & \text{Artificial Intelligence}\\
\textbf{MLM} & \text{Masked Language Modeling}\\
\textbf{CV} & \text{Computer Vision}\\
\textbf{NN} & \text{Neural Network}\\
\end{abbreviations}




\dedicatory{Dedicated to my lovely parents and sister\ldots} 


\mainmatter 

\pagestyle{thesis} 



\chapter{Introduction} 

\label{Chapter1} 


\newcommand{\keyword}[1]{\textbf{#1}}
\newcommand{\tabhead}[1]{\textbf{#1}}
\newcommand{\code}[1]{\texttt{#1}}
\newcommand{\file}[1]{\texttt{\bfseries#1}}
\newcommand{\option}[1]{\texttt{\itshape#1}}


\section{Background}
Deep learning is an extension of ML algorithms known as artificial neural networks, which have made noteworthy advancements in recent years as computational powers have been increased and data has become more generative. This has resulted in improvements in neural network optimization and architectures, which incorporate the deep learning field as we know it today. Deep learning has grown it's popularity in last few years due to its capability to generate cutting-edge artificial intelligence techniques for a large-range of application and data types  \citep{c1}.Nowadays self-supervised deep-neural-networks are heavily employed in image-classification, component detection, language translation, human speech recognition, NLP, and a variety of other applications. Self-supervised learning is currently receiving a fantastic deal of attention in both research and practice. This is possible because deep learning algorithms based on self-supervised learning, which can learn rich patterns from plenty of data obtained from real-world \citep{c2}.
\hfill\\\\
Deep learning is currently receiving a great attention in both AI research and practice. Deep learning has attained popularity because of the incredible value that neural networks have exhibited in areas namely, audio processing, CV, and NLP. The trend for deep learning’s popularity from 2007 to 2022 can be seen in Figure 1.1. 
    \begin{figure}[h!]
        \centering
        \includegraphics[width=0.85\linewidth]{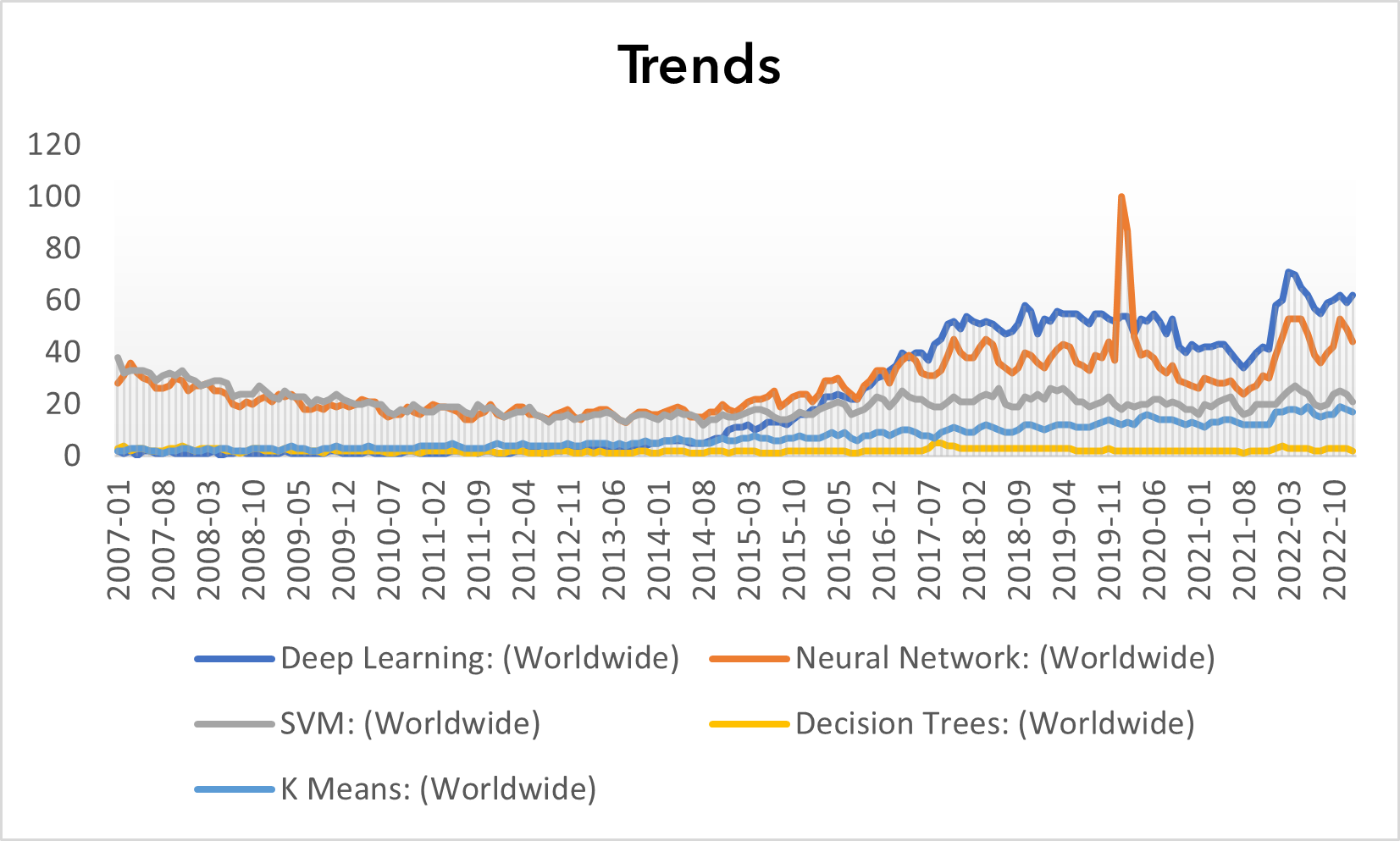}
        \caption{Deep Learning Trend from Google Trends}
        \label{fig:Deep Learning Trend}
    \end{figure}
Additionally, transformer-based deep learning models have demonstrated success in a variety of domains, including image-classification and object identification, as demonstrated by ImageNet, a large, labelled dataset widely used for training and evaluating deep neural networks \citep{c3}. ImageNet has been instrumental in the evolution of transformer-based deep learning computer-vision models, such as the ResNet model, which achieved cutting-edge performance on ImageNet\citep{c4}.
\hfill\\\\
Similarly, the success of transformer based deep-learning natural language processing models is demonstrated by GPT and BERT. These DL models have gained outstanding outcomes on various language understanding tasks, such as speech translation and human-like question answering. ChatGPT and BARD are large language models trained by OpenAI\citep{c5} and Google AI \citep{c6}, respectively, which are based on the GPT and BERT architectures, respectively. Both have demonstrated to be capable of generating human-like text, answering questions, and even carrying out simple conversations.
\hfill\\\\
One of the most promising abilities of self-supervised models is that they can practically simulate any input-to-output relationship in the dataset and generate significant results. This ability has led deep learning to be implemented in wide range of applications. Currently, self supervised learning is applied in various fields. For instance, a deep model was able to generate art (\cite{c13}), it’s been used to improve the pixel quality of images (\cite{c14}), used in autonomous cars for controlling various sections of the car (\cite{c15}), and even recommending movies and shows to the users (\cite{c16}) and various others.
\hfill\\\\
These tools and methods have matured to the point where they can be used to run these systems in a production or commercial setting, resulting in the deployment of several deep learning-based commercial applications such as voice assistants, face recognition, and language translation.
\hfill\\\\
All the above-mentioned deep learning examples have a few things in common, i.e., the values used, or the measurements provided. To put it another way, the data in computer vision represents pixel values, but the data in NLP and audio processing represent words and sound waves. Although it is not a requirement, this may be seen as a factor in the success of deep learning algorithms in several application fields. Since each input characteristic may be handled the same, modeling data made up of the same kind of measurements is easier (\cite{dl_paper}). Furthermore, it is discovered in the aforementioned deep learning applications that there are universal patterns in each of these disciplines. This makes it possible to transfer knowledge between tasks that fall within the same domain.
\hfill\\\\
Both knowledge gained by people and knowledge gained by a deep learning model are included in the knowledge that needs to be communicated (\cite{c17}). For instance, improvements in the classification of pet images in computer vision will probably also make it easier to identify malignancies in X-rays. In other words, a deep learning model may find that the patterns it picks up while working on one task can also be applied to other, related tasks. This phenomenon, which is investigated in the topic of transfer learning, is a second factor in the effective use of deep learning techniques.


\section{Motivation}
However, the common data obtained in the real world is tabular data, which includes samples with sets of features in the form of rows and columns. As compared to image or text data, tabular data is heterogeneous, which has numeric and categorical features that have to be discovered by the neural network to find relations without depending on the spatial data. Therefore, during the last few years, various models and methods have been applied to extract meaningful information from tabular data, and the conventional machine learning models have showcased tremendous performance. A two-dimensional table can be used to represent a tabular dataset, with each row illustrating a single data value type, and each column designating distinct relevant attributes (\cite{c18}). 
\hfill\\\\
Traditional supervised machine-learning techniques, such as Gradient Boosting Decision Tree (GBDT), are widely employed and well-known for their excellent performance in tabular domain applications (\cite{c7}). Some of the popular GBDT algorithms used by researchers and industry practitioners include XGBoost, LightGBM, and CatBoost. GBDT is comprised of a sequence of weak learners, and decision trees which are commonly used in this method. Whereas the XGBoost algorithm is based on the GBDT architecture which has proved to produce excellent results for tabular datasets where it employs a gradient-descent-based algorithm to minimize loss while creating new models from the residuals of previous models (\cite{c7}). Figure 1.2 provides a high-level overview of supervised learning approach.
\hfill\\\\
    \begin{figure}[h!]
        \centering
        \includegraphics[width=0.95\linewidth]{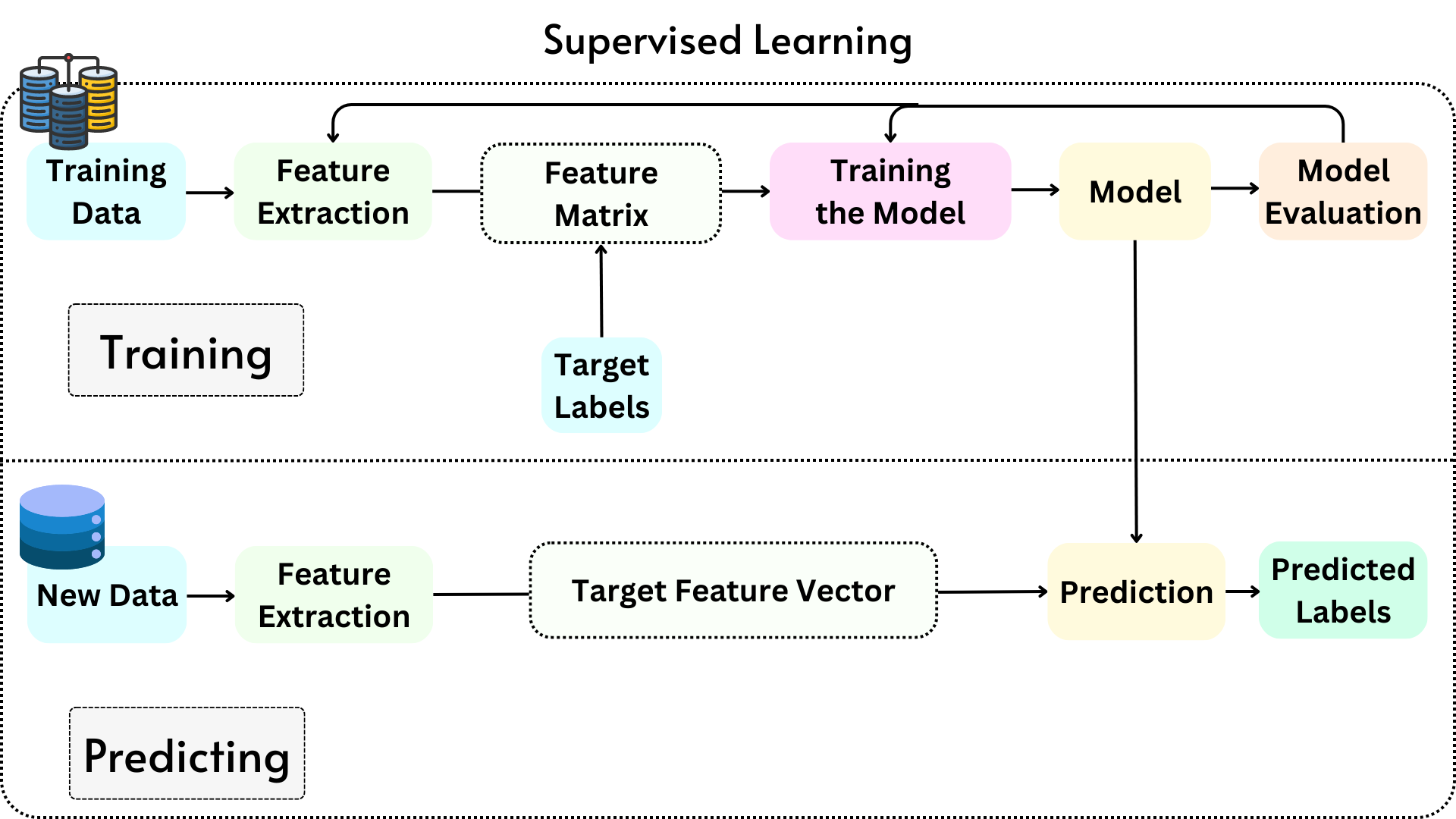}
        \caption{Supervised Learning}
        \label{fig:Supervised Learning}
    \end{figure}
\hfill\\\\
Recently, many researchers have conducted diverse experiments for implementing deep learning on tabular datasets. Research conducted for tabular data falls into various categories: Differentiable decision trees (DDT), this model seeks to differentiate decision trees so that they can be employed as component in end to end train pipelines. This is accomplished by a decision function in the nodes to distinguish tree functioning with it's routing. Attention-based models, enable feature interactions by utilizing attention-like modules where data interact with the samples in each row and column. The regularization method helps to discourage overfitting by adding retribution to the loss function (\cite{atten}). This retribution can be used to automate the tuning of hyperparameters of the model. Another way to design deep learning models for tabular data are integrating feature-products into MLP model. Feature products allow the model to explicitly model multiplicative interactions between features. Lastly, 1D-CNNs can be used to extract the benefits of convolutions networks in tabular data which allows the model to learn local patterns from the dataset. Hence, by observing the recent progress, it might be claimed that the field is nearly as materialized or popular as deep-learning for computer-vision or natural language processing (\cite{cnn_paper}).
\hfill\\\\
Though there are several challenges which needs to be addressed before implementing deep learning on tabular datasets. Some of the challenges are as follows: 1, We must determine how to numerically represent mixed feature data types (categorical and continuous) during model training. 2, Once the optimal representation has been determined, we must devise methods for learning the interactions between the features and how they relate to the target. This is critical for efficient model performance. 3, Tabular datasets are typically smaller than those used in CV and NLP, and there's no large dataset with universal properties available for learning. As a result, we must devise methods to increase sample efficiency. 4, Deep learning is frequently hampered by a perceived lack of interpretability. Thus, we need to find ways to explain model output, enabling its use in a wider scope of applications(\cite{tab_dl_problem}).
\hfill\\\\
Hence, to overcome such challenges we can make use of self supervised learning approach for tabular datasets. Self supervised learning has various advantages when it comes to tabular data. For example, it allows efficient use of unlabeled tabular data, which is frequently abundant and readily available. This eases the training process's dependency on manually labeled data, assembling it more scalable and cost-effective. Hence, by learning meaningful patterns from unlabeled tabular data, self-supervised learning helps to generalize the models application on multiple tasks. Patterns captured allow the model to adapt to different problem domains without the need for task-specific labeled data (\cite{ssl_bert}). Whereas self-supervised learning boosts the robustness and transferability of learned representations, making them more reliable in different scenarios or with limited labeled data. Overall, self-supervised learning is a powerful method for maximizing the potential of unlabeled tabular data, allowing for more efficient and effective learning in deep learning models. 
\hfill\\\\
Now, let’s take a look at how the self-supervised learning is performed. Self- supervised learning refer to idea of training a deep-learning model on a task where it can automatically produce the input and target pairs and passed it to the model, using such intervention we can eliminate the task of data labelling process. In simple words in this process, we want the model to generate data-labels pairs on their own and use their own generated labelled-data for predicting the other section of data(\cite{ssl_vision}). As using SSL, models can be train on a large scale of unlabelled data and then find patterns in the data and provide useful insights on the data. Such process also holds true for tabular data where a dataset is passed to the self-supervised model to predict the other section of the data. Figure 1.3 provides a high-level overview of self-supervised learning approach.
\hfill\\\\
    \begin{figure}[h!]
        \centering
        \includegraphics[width=0.95\linewidth]{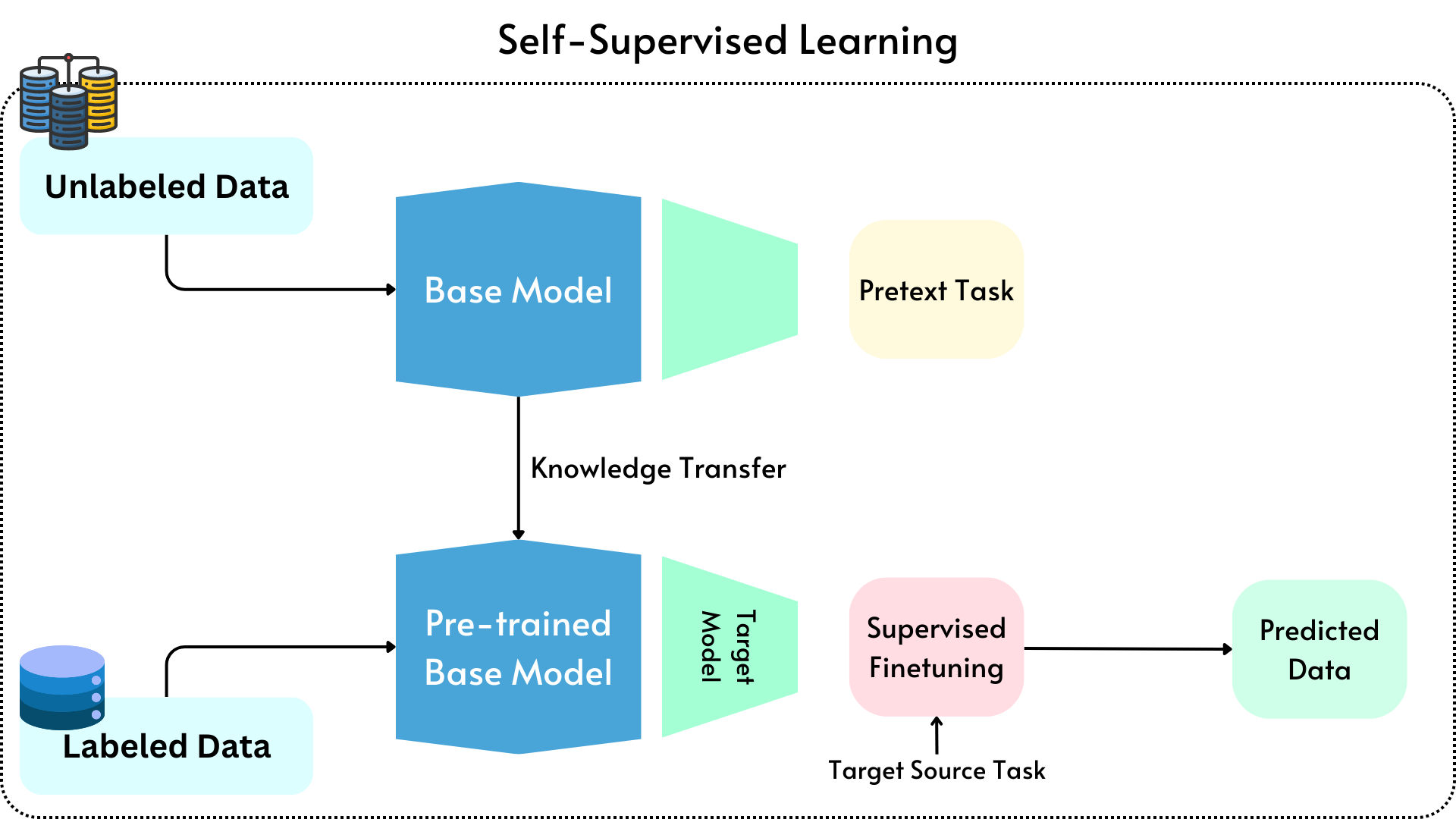}
        \caption{Self-Supervised Learning}
        \label{fig:Self-Supervised Learning}
    \end{figure}
\hfill\\\\
Another advantage of self-supervised learning is that it can help address issues with class imbalance in tabular datasets. In many real-world datasets, the number of instances for each class can be highly imbalanced, which can negatively impact performance of conventional machine-learning model. SSL approaches can be used to generate additional synthetic examples for underrepresented classes, which will help to balance dataset and improve performance of downstream models (\cite{c8}). Additionally, tabular datasets can be one of several input modalities for multimodal learning problems, where self-supervised learning can be used to pretrain models before fine-tuning them on specific downstream tasks. For example, in the field of autonomous driving, self-supervised learning has been used to pretrain models on visual, lidar, and radar sensor data, before fine-tuning the models on specific tasks such as obstacle detection or lane keeping (\cite{c9}).
\hfill\\\\
Hence, self-supervised learning approaches deliver considerable advantages as compared to the conventional ML models. Some of them are as follows – deep neural networks are highly flexible, which helps them to train the models effectively and iteratively (\cite{c10}). They can also help in generation of tabular data as those methods can help address issues with class imbalance. Along with that tabular dataset can be one of several inputs modalitie for multimodal learning issues where self-supervised learning can be used (\cite{c8}).

\section{Objective}
There are several methods for training tabular data, including using traditional machine learning models such as GBDT or implementing supervised or unsupervised deep learning methodologies. The core aim of our research is to train tabular data on TabTransformer using self-supervised learning approach, in which a model learns from unlabeled data by creating surrogate supervised tasks, rather than relying on labelled data. The TabTransformer is a Transformer-based model optimised for tabular data. The Transformers architecture, which was mainly designed for NLP tasks, is been adapted for tabular dataset by including appropriate input embeddings for categorical and numerical values with positional encodings. The TabTransformer uses Transformers' self-attention mechanism to capture relationships and dependencies between features in tabular data. 
\hfill\\\\
As a result, we aim to find the most effective representation of categorical and numerical features for the TabTransformer model by experimenting with various input settings, which will allow us to construct categorical and numerical inputs for the Transformers and overcome the challenges mentioned in the motivation section. And finally, compare the TabTransformer model's performance to baseline models such as MLP (Multi-Layer Perceptron) and Supervised TabTransformer. This comparison will show how the Self-supervised TabTransformer perform against the traditional machine learning models and other supervised learning approaches.
				
\chapter{Literature Review}
\label{chap2}

As per the research, there are not much dedicated research paper which examines the recent self-supervised deep learning models on tabular data and evaluate their performance. Though, there are several studies which includes the state-of-th-art model GBDT with comparison of various deep supervised learning models like ResNet, MLP, TabNet, AutoInt and several others. For tabular datasets, decision tree ensembles like GBDT (\cite{c11}) and random forests (\cite{c12}) are currently the best options for the researchers for tabular data. Many prominent GBDT variants are currently in use by academics and ML practitioners which includes "XGBoost (\cite{c7}), LightGBM  (\cite{c19}), and CatBoost  (\cite{c20})". The main disadvantage of tree-based systems is that they frequently forbid end to end optimisation and make use of greedy, local optimisation techniques to build trees. As a result, they cannot be trained as an end-to-end pipeline component. In-order to solve this problem, various publications (\cite{c21}) and (\cite{c22}) and (\cite{c23}) suggest "softening" in the decision-making processes to the internal tree node in order to differentiate the general tree functioning from trees routeing.
\hfill\\\\
A recent survey by Gorishniy et al.  (\cite{c9}), have evaluated and compared the results of several deep learning models and have trained it on various tabular data set, along with that they have also proposed a novel Transformer architecture named as FT (feature tokenizer)-Transformer which turns all the features into tokens and run the generated stacks into the transformer layer. Levin et al. (\cite{c28}) have performed transform learning on a medical tabular dataset, on which various supervised and self-supervised models have been examined and compared the models with GBDT based – Catboost and XGBoost. Recently, Gorishniy et al. (\cite{c29}) have examined many deep learning models on a large range of dataset and have evaluated the results. Their findings states that a tuned ResNet model is able to perform well as compared to some other optimal deep learning models.
\hfill\\\\
Shwartz and Armon have (\cite{c30}) evaluated various deep learning models like TabNet (\cite{c31}), Net-DNF (\cite{c32}), NODE (\cite{c33}) and have compared the deep approaches of trees on accuracy, efficiency, parameter optimization and various others. Numerous papers (\cite{c24}) and (\cite{c25}) and (\cite{c26}) in the publications on RS and the projection of click through rates criticize MLP because of its inadequacy in simulating multiplicative interchanges between characteristics. This inspiration served as the motivation for some works [27, 29] that suggested various approaches to integrating feature items into MLP.
\hfill\\\\
"BERT-Bidirectional Encoder Representations from Transformers"  (\cite{c1}) has been successful in a variety of applications involving natural language processing. By applying conditions on both the left and the right context in all Transformer levels, BERT is able to pre-train deep bidirectional representation. In our study we will be considering the masking process from the BERT to apply in our modelled dataset. The foundation of The TabTransformer (\cite{c27}) is self-attention-based Transformers. In order to increase prediction accuracy, the Transformer layers convert the embedded categorical characteristics into robust contextual embedding. Thus, various authors have proposed numerous models for examining tabular dataset.
\hfill\\\\
(\cite{yoon}) proposed a method for tabular data that combines a denoising auto-encoder with a classifier connected to its representational layer. Encoder receives corrupted input data from a random binary mask network. While the decoder, like a denoising autoencoder, attempts to reconstruct the original uncorrupted input, the classifier predicts the mask. This approach, may not be suitable for very high dimensional or smaller, and noisy-datasets, as the model can easily become over parameter tuning and perform overfitting. 
\hfill\\\\
Similarly, training a classifier in this configuration is difficult because it must predict highly dimensional, sparses values, and unbalanced binary masked values, which is similar to difficulties encountered by training models on unbalanced binary datasets (\cite{ucar}). (\cite{verma}), on the other hand, used "mixup", a technique that generates positive pairs for InfoNCE loss by merging pairs of data and representational layers. They saw improvements in a variety of domains, including a tabular data model where image data was flattened and permuted.
\hfill\\\\
Self-supervised learning is a learning method that does not necessitate labelled data. Instead, by completing a pretext task, the model learns from unlabeled data. Pretext tasks are intended to encourage the model to learn useful data representations. Existing work on SSL that can applied to tabular datasets exists. The pre-text tasks in denoising autoencoders is by recovering original sample from a corrupted sample. The pre-text task in context encoders is to reconstruct the original sample from both the corrupted sample and the mask vector. In TabNet and TaBERT, the pre-text task for SSL is also recovering corrupted tabular data (\cite{survey}).
Here is a table that summarizes the different self-supervised learning approaches for tabular data:
    \begin{figure}[h!]
        \centering
        \includegraphics[width=0.70\linewidth]{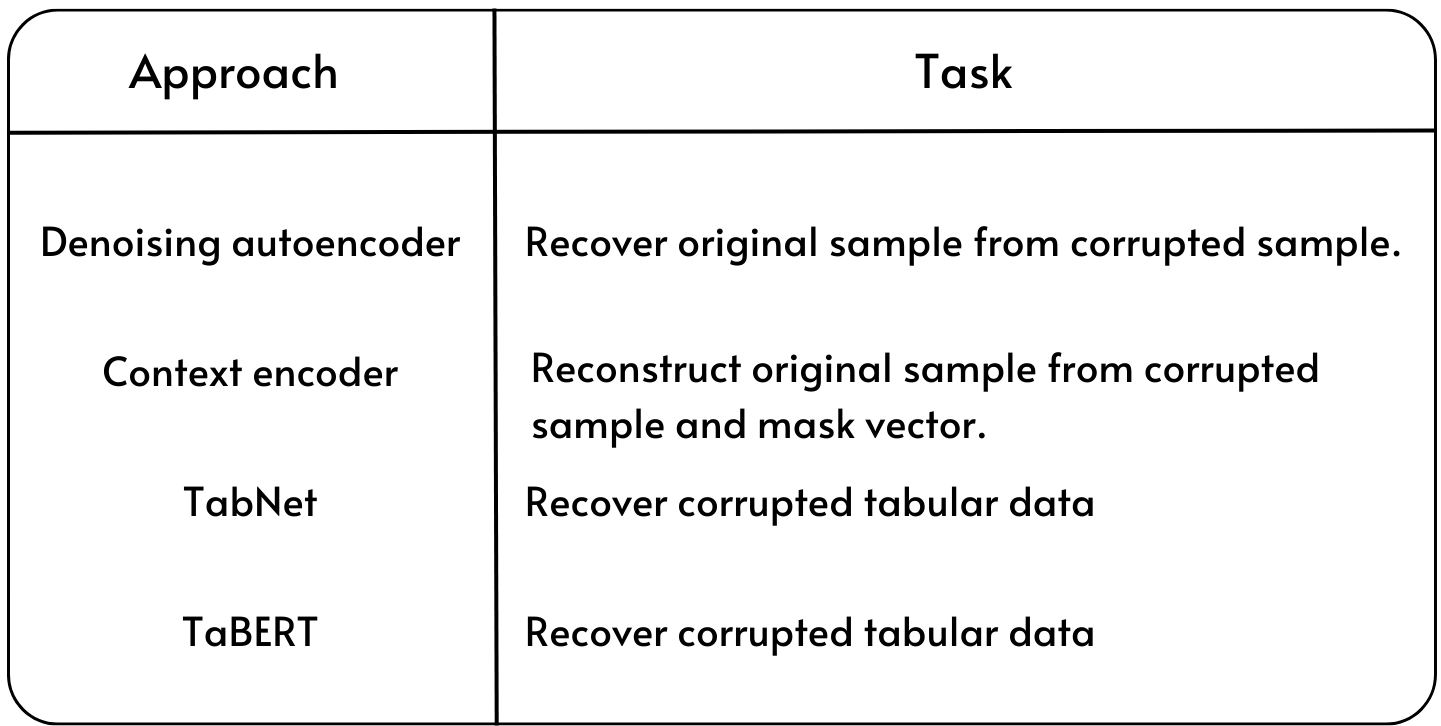}
        \caption{Self-supervised learning approaches}
        \label{fig:Self-supervised learning approaches}
    \end{figure}
\hfill\\\\   
However, there were few innovative approaches observed such as (\cite{ucar}) proposed SubTab, a new architecture for learning from tabular data. SubTab separates the input features into subsets before re-constructing the data from each subset. According to the authors, this method effectively captures the underlying latent representation of the data. By combining the latent representations of the subsets, the SubTab framework learns a mutual representation.
\hfill\\\\
However, the use of contrastive and distance losses introduces computational complexity because it necessitates projection combinations. Though, The SubTab's reconstruction step is unsuitable for high-dimensional data, which is common in many life science applications. SCARF (\cite{bahri}) is another model that is based on contrastive learning and extends existing inputs in structured domains. For a given input, SCARF generates negative pairs by selecting a random subset of features and replacing them with random draws from the corresponding observed borderline allocations. Hence, various authors have proposed numerous models for examining tabular dataset.

\chapter{Methodology}
\label{chap3}

\section{Methodology Background}

In this chapter, we will discuss various \textbf{TabTransformer Variants} and explore how \textbf{Self-supervised learning} is used to improve TabTransformers for tabular data.
\hfill\\\\
The TabTransformer is a variant of the Transformer architecture which is specifically designed for handling tabular data(\cite{c27}). In order to perform efficient modelling and prediction tasks, it aims to capture dependencies and relationships between features in a table. The encoder-decoder architecture of the TabTransformer is comparable to that of the first Transformer model. In SSL, a model learn from data on its own without the need for explicit human-labeled annotations. Instead, the model is conditioned to anticipate specific attributes or traits of data itself. SSL can be used in the context of tabular data to derive useful representations or embeddings from the input features.
\hfill\\\\
Various TabTransformer variants are used to pretrain the model on tabular data using self-supervised learning techniques which are discussed in next section. Performing such task the model learns meaningful representations of the input features during this pretraining phase and the knowledge gained by the model can be used for vaiours tasks like classification, regression, or anomaly detection.
\hfill\\\\
There are several self-supervised learning techniques that which can be used on TabTransformers for tabular data. Some of the commonly used techniques are as follows:
\hfill\\\\
1.	Contrastive learning: Contrastive learning is a type of SSL approach where the model learns to identify between positive and negative samples. Positive samples are pairs of similar instances, while negative samples are pairs of dissimilar instances. The model learns to do this by maximizing the similarity of the positive samples and minimizing similarity of the negative samples.Hence, model can capture meaningful representations by learning to distinguish between similar and dissimilar values in the dataset which can be later used for prediction(\cite{cont_learn}).
\hfill\\\\
2.	Masked Language Modelling Approach (MLM): Using Masking technique, some of input feature are randomly masked, and the model is trained to forecast or predict original values based on context of other features used while training. The TabTransformer can detect and learn the dependencies and relationships between features while recovering those features(\cite{mlm_learn}).
\hfill\\\\
3.	Autoencoding: Using a compressed or encoded representations model is trained by recreating original input features. Here model develops the ability to extract most significant data of the input data by squeezing and later reconstructing the features(\cite{autoenco}).
\hfill\\\\
I have chosen MLM Approach for our research and experiments to perform self-supervised learning on tabular data.
\hfill\\\\
The TabTransformer can learn complex tabular data representations using these self-supervised learning techniques without labelled data. The pretrained TabTransformer can be fine-tuned on labelled data after the pretraining phase is over for particular downstream tasks, like classification or regression(\cite{c27}). Therefore, we can efficiently use sizable amounts of unlabeled tabular data to improve the performance and generalisation of models on various tabular data tasks by utilising self-supervised learning on the TabTransformer architecture.

\section{Transformer}
In this section, we will begin by understanding the Transformer, which serves as the underlying architecture for the TabTransformer. Subsequently in next section, we will dive into multiple variants of the TabTransformer that I have developed specifically for tabular data. These variants incorporate enhancements and modifications to the original TabTransformer to improve its performance and suitability for handling tabular datasets. Now, let's dive into what transformer's are.
\hfill\\\\
The most recent advancements in NLP and CV systems heavily depend on the use of transformers. However, gradient-boosted decision trees (GBDT) continue to be widely used for tabular data analysis. Hence, to fill this gap (\cite{c27}) first described the transformer-based tabular data modeling models in their study titled "TabTransformer: Tabular Data Modelling Using Contextual Embeddings." Before going in-depth into how TabTransformer’s work we will understand how the Transformer works.
\hfill\\\\
The "Attention Is All You Need" paper by (\cite{transformer})introduces the Transformer, a brand-new neural network architecture. Transformer is focused on changing one sequence into another which can be comparable to LSTM models. To learn long-range dependencies between input and output sequences, the Transformer makes use of an attention mechanism where it can focus on particular segments of the input sequence when creating the output sequence, that's how the attention mechanism operates. Hence, to understand the core architecture of the Transformer let’s take a look at its architecture as shown below:
\hfill\\\\
The Transformer model contains of mainly two key components: the encoder and the decoder. Both parts are being constructed using modules that can be stacked on top of one another and these modules include layers for feed-forward and multi-head attention. The input and output sentences are first transformed into a higher-dimensional space before processing. The positional encoding of individual value is an essential component of the Transformer model. The embedded representation of each value includes positional information to get around this restriction.Which helps to learn the positional relationships between the value in the sequence model(\cite{transformer}).
    \begin{figure}[h!]
        \centering
        \includegraphics[width=0.63\linewidth]{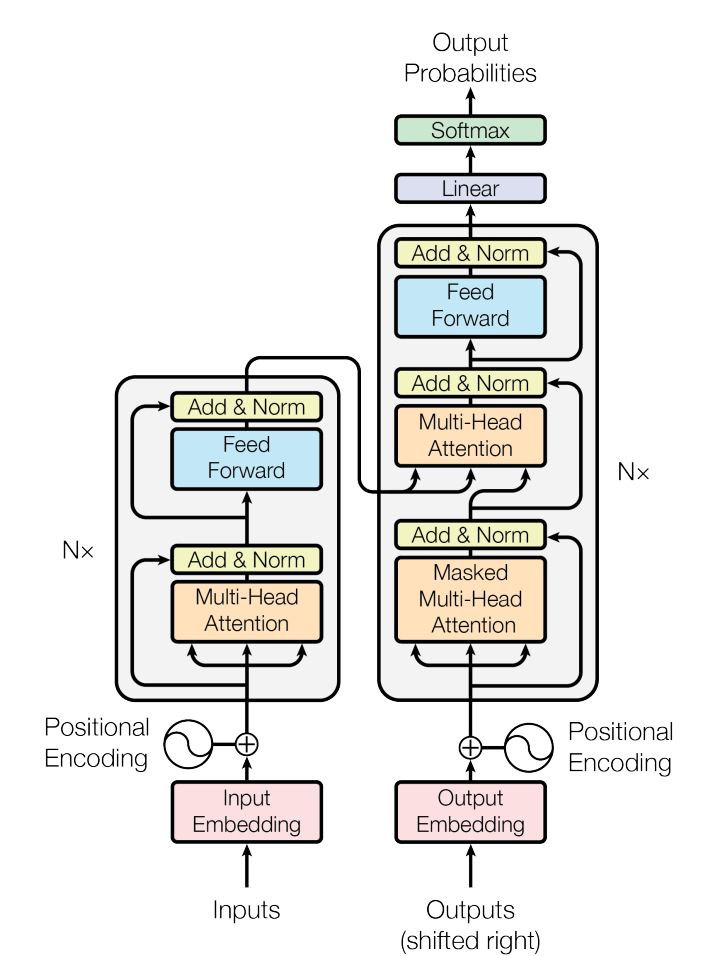}
        \caption{Transformer Architecture by authors (\cite{transformer})}
        \label{fig:Transformer Architecture by authors}
    \end{figure}
\hfill\\\\
Now, let's dive into a more detailed explanation of each component:
\hfill\\\\  
\textbf{Encoder}: Here the encoder's role is to take a sequence of input data and turn it into a sequence of hidden representations. Which is accomplished by using a stack of "encoder blocks where multi-head-attention layers and feed-forward layers are used in each encoders block. The model can learn long-range dependencies between words thanks to the multi-head attention layer's ability to concentrate on various input sequence segments"(\cite{transformer}). On the other hand, the feed-forward layer is in charge of discovering non-linear connections between the input and output representations. The Encoder architecture can be seen in Figure 3.2.
    \begin{figure}[h!]
        \centering
        \includegraphics[width=0.28\linewidth]{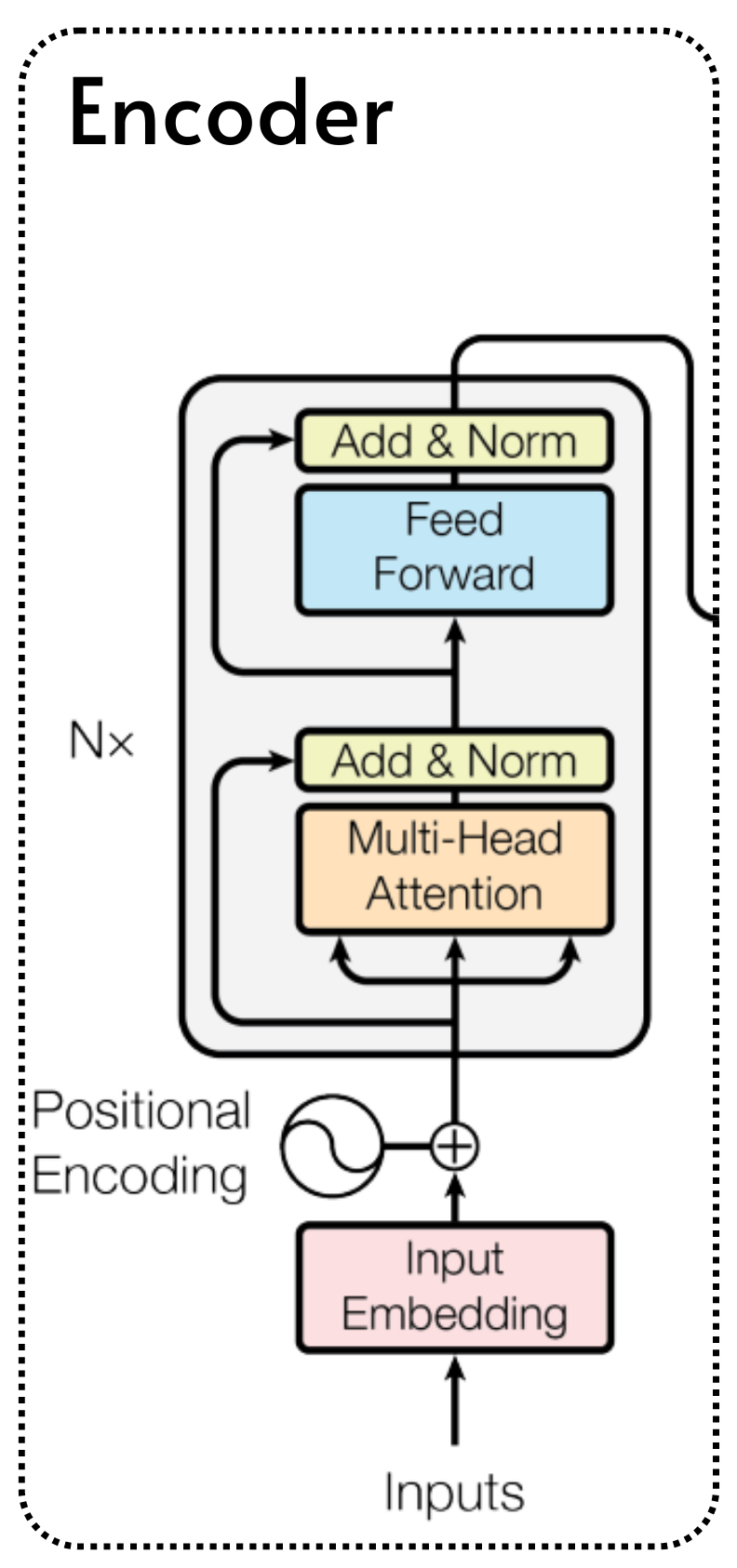}
        \caption{Transformer's Encoder Architecture}
        \label{fig:Transformer's Encoder Architecture}
    \end{figure}
\hfill\\\\
\textbf{Decoder}: After receiving the encoder's hidden representations, the decoder component creates the output sequence. It is made up of a stack of decoder blocks, each of which has a residual connection, feed-forward-layer, with multi-head-attention layer. "Similar to encoder, decoder's multi-head-attention layers helps model to focus with various segments of the input sequence"(\cite{transformer}). When gradients pass through the network, the residual connection helps to protect crucial information while the feed-forward layer learns non-linear relationships between the representations. pear connections between the input and output representations. The Decoder architecture can be seen in Figure 3.3.
    \begin{figure}[h!]
        \centering
        \includegraphics[width=0.30\linewidth]{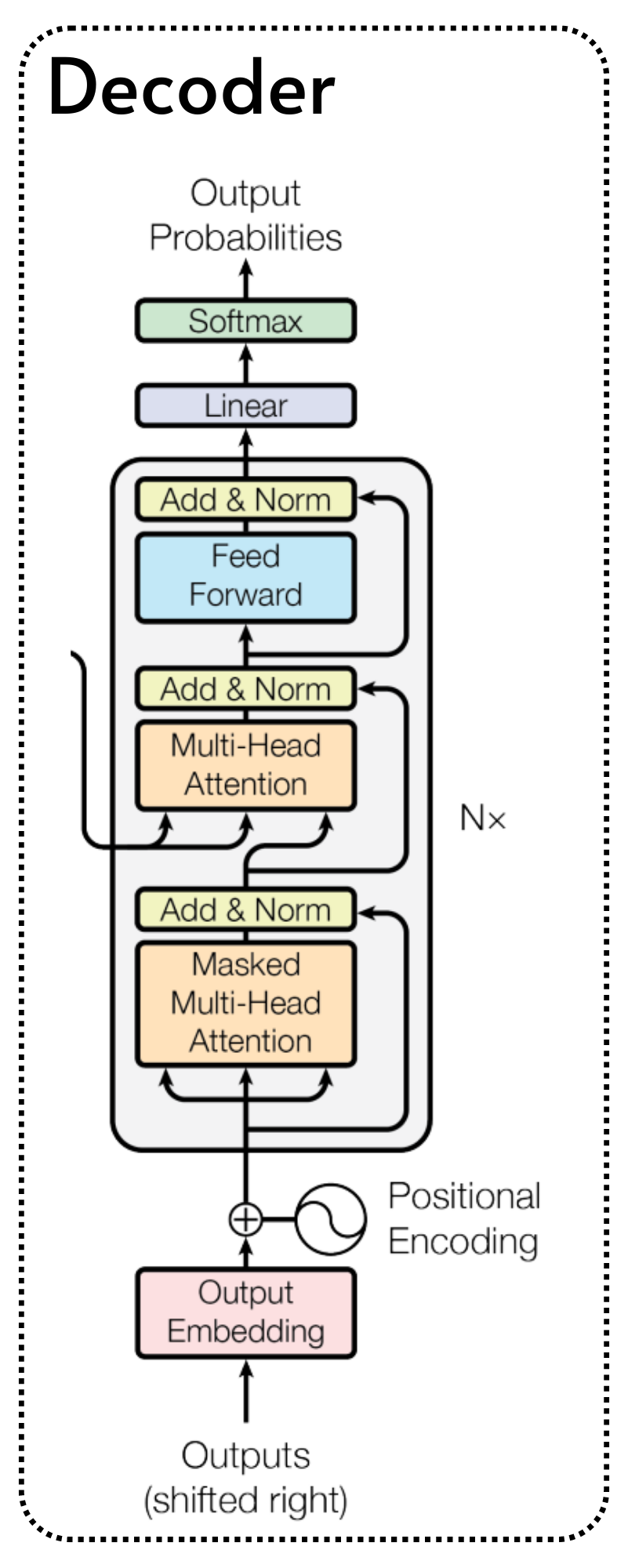}
        \caption{Transformer's Decoder Architecture}
        \label{fig:Transformer's Decoder Architecture}
    \end{figure}
\hfill\\\\
\textbf{Multi-head attention}: Multi-head attention is approach which allows model to concentrate on different elements of input sequence at once. This is accomplished by carrying out numerous attention computations concurrently(\cite{transformer}). The model can better understand the input by capturing diverse dependencies and paying attention to different positions in the sequence.
    \begin{figure}[h!]
        \centering
        \includegraphics[width=0.85\linewidth]{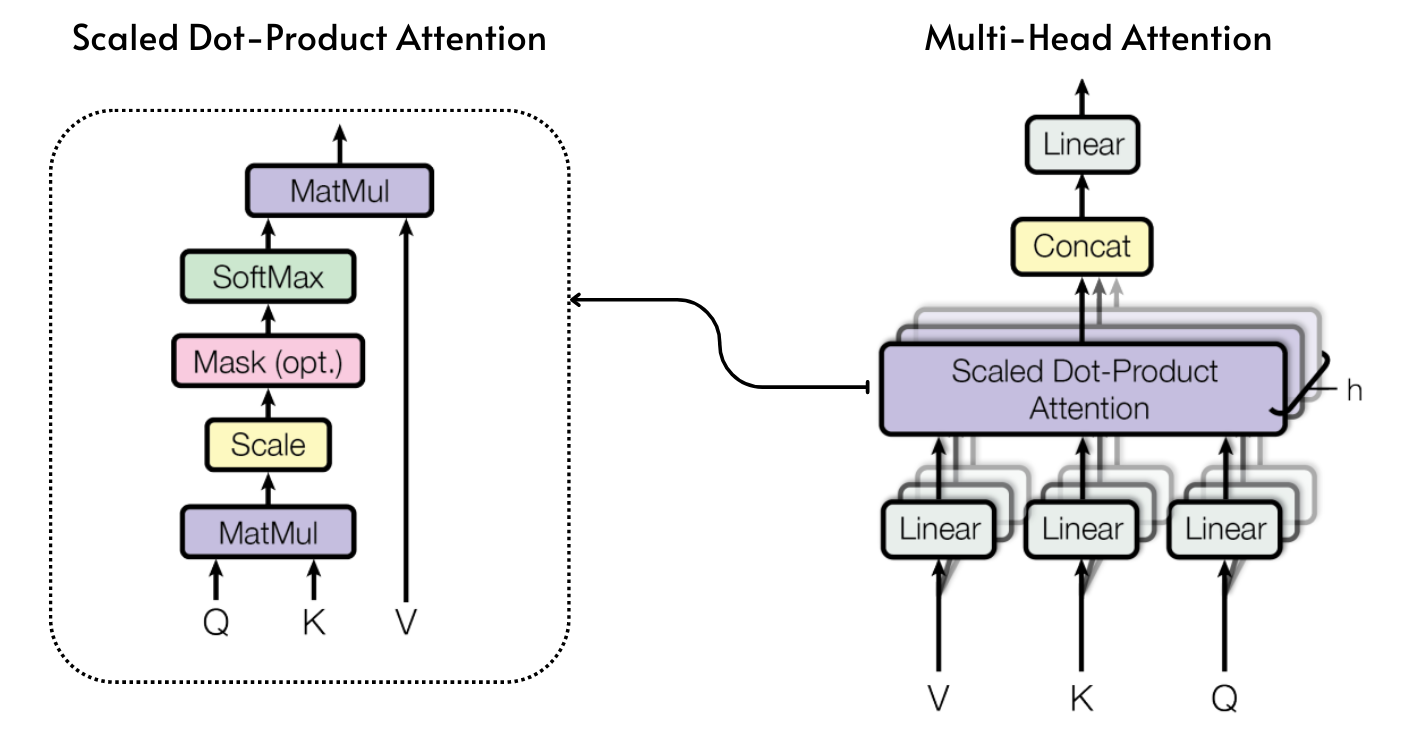}
        \caption{Attention Mechanism}
        \label{fig:Attention Mechanism}
    \end{figure}
\hfill\\\\
Here in multi-head attention, there are 3 matrices— 1.query matrix(Q), 2.key matrix(K), and 3.value matrix(V) as shown in equation 3.1—representing word sequences, values in V are multiplied and added with attention-weights (a), which establishes the impact of each and every words in sequence (Q) with all other words in sequence(K)(\cite{transformer}). 
\hfill\\\\
\begin{equation}
\text{Attention}(Q, K, V) = \text{softmax}\left(\frac{Q K^T}{\sqrt{d_k}}\right) V
\end{equation}
\hfill\\\\
To create a distribution between 0 and 1, the weights are subjected to the SoftMax function. The model learns from various representations of Q, K, and V by parallelizing attention mechanism into multiple mechanisms. Using learned weight matrices (W), it is possible to obtain linear projections of Q, K, and V. Matrice-Q, K, \& V vary according to the position of the attention module in encoder-decoder, or else between encoder-decoder(\cite{transformer}). Following the attention heads, there is a pointwise feed-forward-layers which applies identical transformations to each element of the sequence.
\hfill\\\\
\textbf{Feed-forward layer}: The feed-forward layer adds non-linearity to the model through a straightforward linear transformation. Applying a series of operations like matrix multiplication and non-linear activation functions enables model to learn intricate connections of the input-output representations(\cite{transformer}).
\hfill\\\\ 
\textbf{Positional encoding}: Positional encoding is used to provide information about the word positions since the Transformer model lacks recurrent networks that naturally learns order of words of a sequence. Using that model is able to recognize the arrangement of words in the output sequence and successfully learn the relationships between them thanks to the addition of this encoding to each word's embedded representation(\cite{transformer}).
\hfill\\\\ 
\section{TabTransformer Variants}
Now, let's explore various variants of TabTransformer.
\subsection{Vanilla – TabTransformer (Vanilla -TT)}

Here Vanilla refers to the original TabTransformer proposed by the authors.
\hfill\\\\  
The main idea behind the TabTransformer is to combine Transformers and Multi-layer Perceptron (MLP) to improve the handling of categorical features. In traditional approaches, categorical values are represented using one-hot encoding or embeddings, which may not capture contextual information and dependencies between categories which can limit the MLP's ability to learn from categorical features effectively. 
\hfill\\\\  
To solve this problem, the TabTransformer author suggests converting the categorical embeddings into contextual embeddings using Transformers(\cite{tabt}). Transformers are well-known for their aptitude at identifying dependencies and connections in sequential data. Contextual embeddings are able to capture interactions between categories and offer a richer representation of the categorical features by utilising the self-attention mechanism of Transformers.
\hfill\\\\  
In this process, the Transformer model - gives categories weights based on how they relate to one another which is feded to the regular categorical embeddings. The transformation of embeddings is incorporated by the attention-based weighting mechanism which allows capturing pertinent contextual information. For the categorical features, the transformed embeddings provide better contextual information.
\hfill\\\\  
The MLP architecture then incorporates these modified contextual embeddings so that they can be used alongside numerical or continuous features. Thus MLP can predict outcomes with greater accuracy by leveraging the improved categorical feature representation. Hence to sum-up, the TabTransformer fuses Transformers and MLP to convert standard categorical embeddings into contextual ones. Due to the model's ability to incorporate dependencies and contextual data, the MLP performs better when learning from categorical features and producing more precise predictions.
\hfill\\\\  
Now, let’s understand the architecture of TabTransformer, architecture can be seen in Figure 3.5.
\hfill\\\\ 
Let's breakdown the architecture of TabTransformer into five sections:
\hfill\\\\
1. \textbf{Numerical or Continuous Feature Normalization}: The first step in the TabTransformer architecture involves normalizing the numerical features. This normalization process ensures that the numerical features have a consistent scale and range which can improve the learning process.
\hfill\\\\
2. \textbf{Categorical Feature Embedding}: The categorical features are converted from scalar to dense vectors representations which are often stated as embeddings. Embeddings capture the semantic meaning and relationships between different categories. This process helps in converting the categorical features into a continuous representation which can be processed by the Transformer.
\hfill\\\\
    \begin{figure}[h!]
        \centering
        \includegraphics[width=0.75\linewidth]{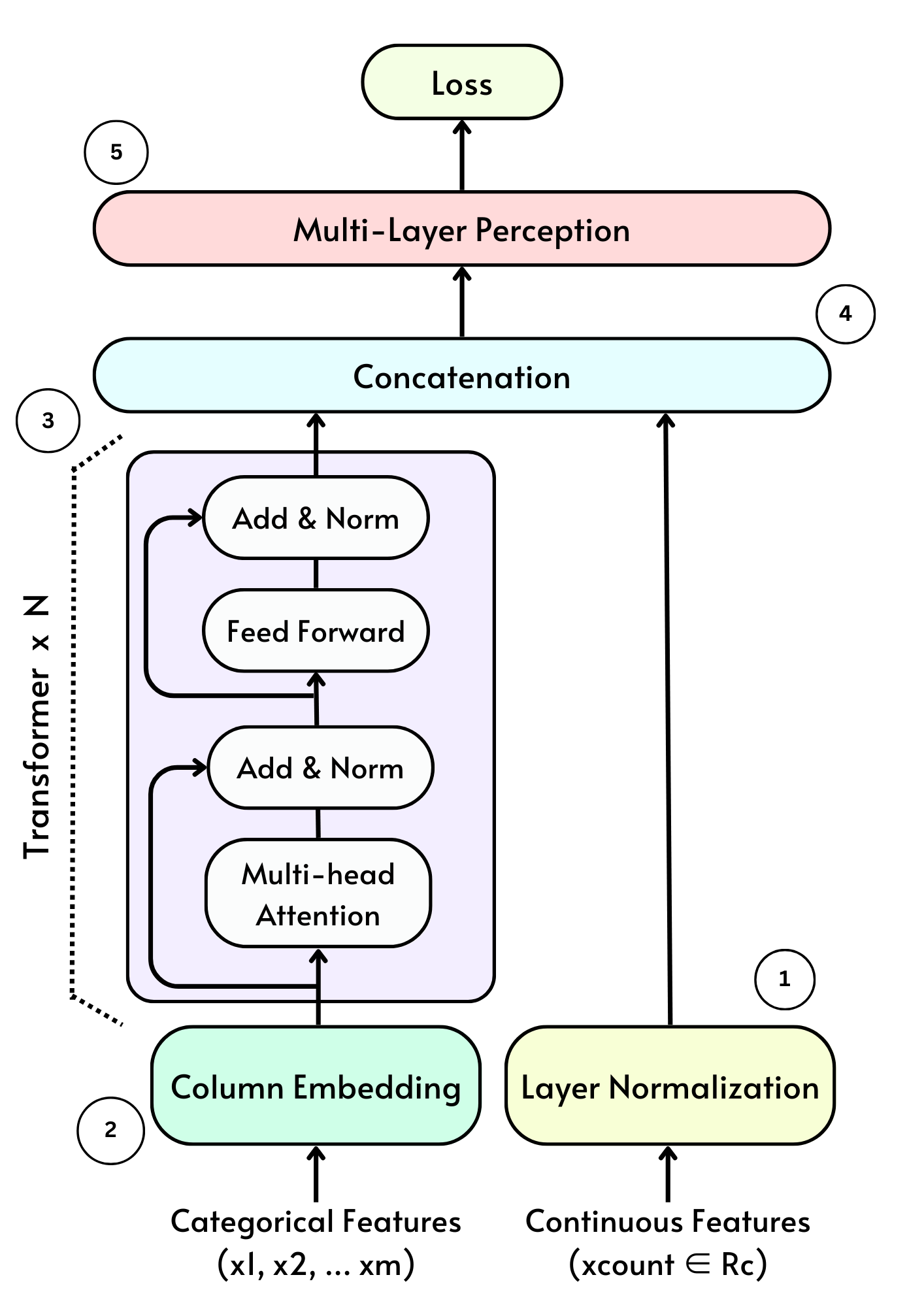}
        \caption{Vanilla TabTransformer Architecture}
        \label{fig:Vanilla TabTransformer Architecture}
    \end{figure}
\hfill\\\\
3. \textbf{Transformer Blocks for Contextual Embeddings}: The embedded categorical features are then passed through multiple Transformer blocks, typically denoted as N as shown in figure. Each Transformer block consists a self-attention-layers and feed-forward NN. Self-attention-mechanism allows model to capture dependencies and relationships between different categorical features, generating contextual embeddings. The Transformer blocks are stacked to capture increasingly complex relationships and dependencies. In-depth explanation of the Transformer is mentioned in section 3.2.
\hfill\\\\
4. \textbf{Concatenation of Contextual Categorical and Numerical Features}: After obtaining the contextual embeddings from the Transformer blocks, they are concatenated or combined with the normalized numerical features. This concatenation combines the improved contextual information from the categorical embeddings with the numerical features, which allows the model to utilize both types of knowledge for making predictions.
\hfill\\\\
5. \textbf{MLP for Prediction}: The concatenated features are then passed to a MLP architecture. Where the MLP consists of multiple layers of interconnected neurons that can learn complex relationships and patterns in the concatenated feature representation. MLP layer processes the concatenated features to generate the final predictions for the desired task like classification or regression.
\hfill\\\\
Hence to summarise, Vanilla-TabTransformer architecture involves normalizing the numerical features and perform embedding of the categorical features by passing the embeddings through Transformer blocks to obtain contextual embeddings of each feature and further concatenating the contextual categorical embeddings with the numerical features and finally using an MLP to make the final predictions based on problem and training the model with the associated loss functions. A detailed visual summary of the Vanilla-TT is mentioned in Appendix.

\subsection{Binned – TabTransformer (Binned -TT)}

Here Binned refers to the binning process performed on numerical values to convert them into categorical values so we can pass those values to embedding layer of the TabTransformer.
\hfill\\\\
The Binned-TT combines Transformers and Multi-layer Perceptron (MLP) to improve the handling of categorical and continuous features. In the Vanilla-TT, the numerical values were not passed to the Transformer which results into the lower accuracy of the Vanilla-TT as it was not having the contextual information of the numerical values. Hence to address this limitation, the Binned-TabTransformer introduces a process where numerical values are binned and converted into categorical values. 
\hfill\\\\
The binning process divides the numerical values into discrete intervals or bins, effectively converting them into categorical representations. These new categorical representations are combined with the remaining categorical values which helps the embedding to treat them as categorical features.
\hfill\\\\
The Binned-TT then utilizes Transformers to transform these combined categorical values into contextual embeddings. Transformers excel at capturing dependencies and relationships in sequential data, making them suitable for capturing the contextual information in the categorical embeddings. By utilizing the self-attention mechanism of Transformers, the model assigns weights to different categories based on their relationships with other categories in the sequence. This attention-based weighting mechanism helps the model focus on relevant contextual information during the embedding transformation process.
\hfill\\\\
The transformed contextual embeddings are then passed into the MLP architecture. Therefore, now MLP can leverage these improved categorical representations of the numerical values. By employing the improved contextual information from the categorical embeddings, the MLP is able to make more accurate predictions.
\hfill\\\\
Now, let’s understand the architecture of Binned-TabTransformer, architecture can be seen below:
\hfill\\\\
    \begin{figure}[h!]
        \centering
        \includegraphics[width=0.70\linewidth]{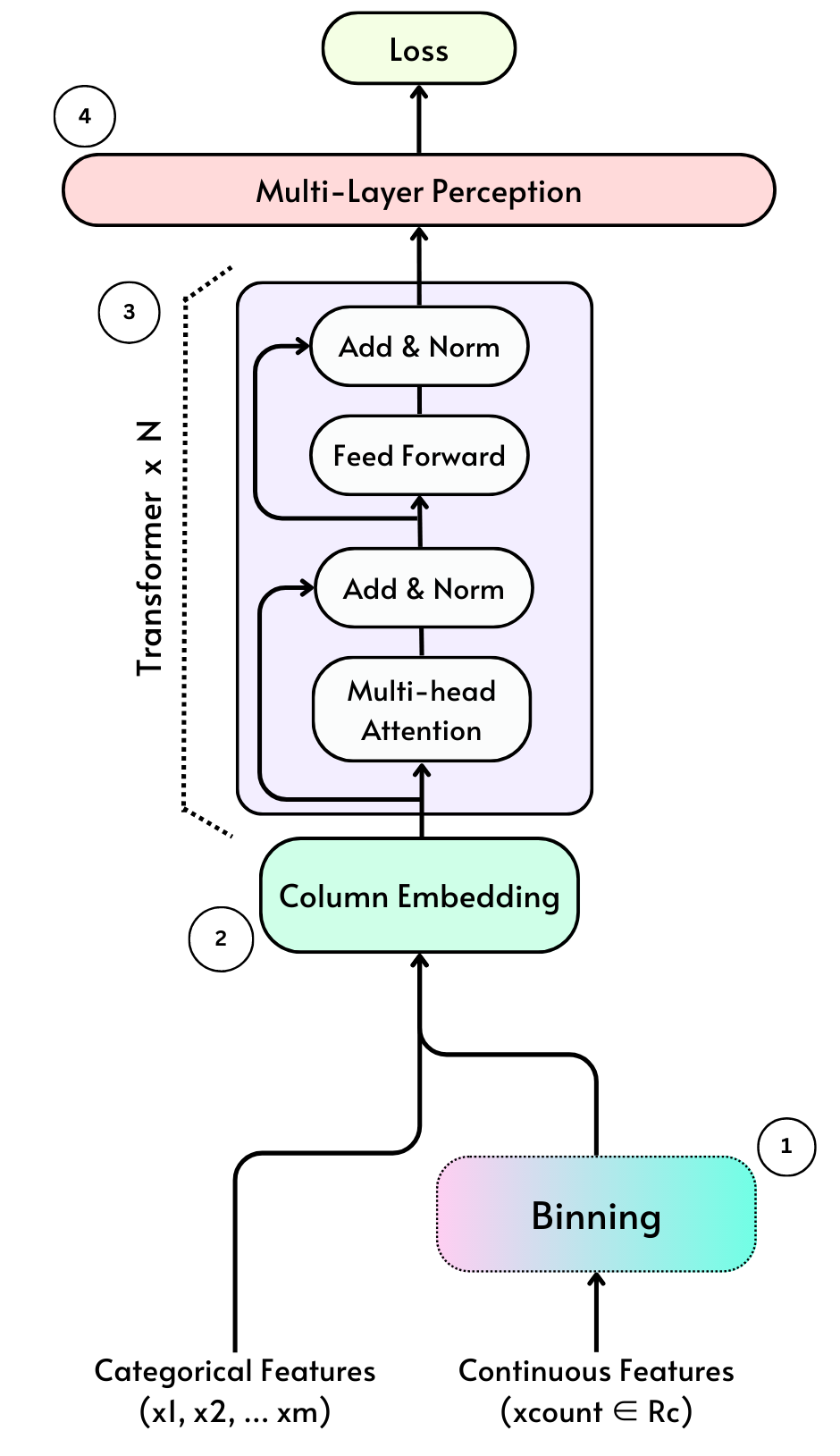}
        \caption{Binned TabTransformer Architecture}
        \label{fig:Binned TabTransformer Architecture}
    \end{figure}
\hfill\\\\
Let's breakdown the architecture of Binned-TabTransformer into four sections:
\hfill\\\\
1. \textbf{Numerical or Continuous Feature Binnig}: The first step in the Binned-TabTransformer architecture involves binning the numerical features. This binning process ensures that the numerical features have been converted to the categorical features, which will allow to fuse the features into the Transformer.
\hfill\\\\
2. \textbf{Categorical and Binned Numerical Feature Embedding}: Now the numerical features will be treated as categorical features. Thus, the combined features will be converted from scalar to dense vectors representations which are often stated as embeddings. Embeddings capture the semantic meaning and relationships between different categories. This process helps in converting the combined categorical features into a continuous representation which can be processed by the Transformer.
\hfill\\\\
3. \textbf{Transformer Blocks for Contextual Embeddings}: The embedded combined categorical features are then passed through multiple Transformer blocks, typically denoted as N as shown in figure. Each Transformer block consists of self-attention layers and feed-forward NN. The self-attention mechanism allow model to capture dependencies and relationships between different categorical features, generating contextual embeddings. The Transformer blocks are stacked to capture increasingly complex relationships and dependencies.
\hfill\\\\
4. \textbf{MLP for Prediction}: The embedded feature from Transformer are passed to a Multi-layer Perceptron architecture. Where the MLP contains of multiple layers with interconnected neurons that can learn complex relationships and patterns in the concatenated feature representation. MLP layer processes the concatenated features to generate the final predictions for the desired task like classification or regression.
\hfill\\\\
Hence to summarize, the Binned-TabTransformer makes use of the binning process to convert numerical values into categorical representations. The Transformers then transform these categorical representations into contextual embeddings, capturing dependencies and contextual information. The MLP incorporates these enhanced categorical embeddings with the numerical features, to improve the model's ability to learn from categorical features and make more precise predictions. A detailed visual summary of the Binned-TT is mentioned in Appendix.

\subsection{Vanilla-MLP – TabTransformer (VM-TT)}

Here Vanilla-MLP refers to the conversion process performed on numerical by a MLP model for converting the scalar numerical values to dense vectors before concatenating them with the contextual embeddings.
\hfill\\\\
The Vanilla-MLP-TT combines Transformers and Multi-layer Perceptron (MLP) to improve the handling of categorical and numerical features. In the Vanilla-TT, the numerical values were passed to MLP layer of the TabTransformer which results in the lower accuracy of the Vanilla-TT as it was not having the dense information of the numerical values. Hence to address this limitation, the Vanilla-MLP-TabTransformer introduces a process where numerical values are passed to an MLP for converting the scalar values to dense before concatenating them with the contextual embedding.
\hfill\\\\
In this process, the Vanilla-MLP-TT model’s Transformer assigns weights to categories based on their relationships with one another, which are then applied to the regular categorical embeddings same as the Vanilla-TT. The transformation of embeddings incorporates an attention-based weighting mechanism, allowing the model to capture relevant contextual information. These transformed embeddings provide enhanced contextual information for the categorical features. 
\hfill\\\\
Whereas for numerical values, the Vanilla-MLP-TT uses an MLP architecture to convert the numerical values into dense vectors before concatenating them with the contextual embeddings. The MLP processes the numerical values by transforming them into continuous representations that capture the numerical information more effectively. These dense vectors are then concatenated with the contextual embeddings from the Transformer allowing the MLP model to leverage both the enhanced categorical representations and the transformed numerical dense vectors. Hence to sum-up, the Vanilla-MLP TabTransformer fuses Transformers and MLP to convert standard categorical embeddings into contextual and numeric to dense vectors.
\hfill\\\\
Now, let’s understand the architecture of Vanilla-MLP-TabTransformer, architecture can be seen below:
\hfill\\\\
    \begin{figure}[h!]
        \centering
        \includegraphics[width=0.70\linewidth]{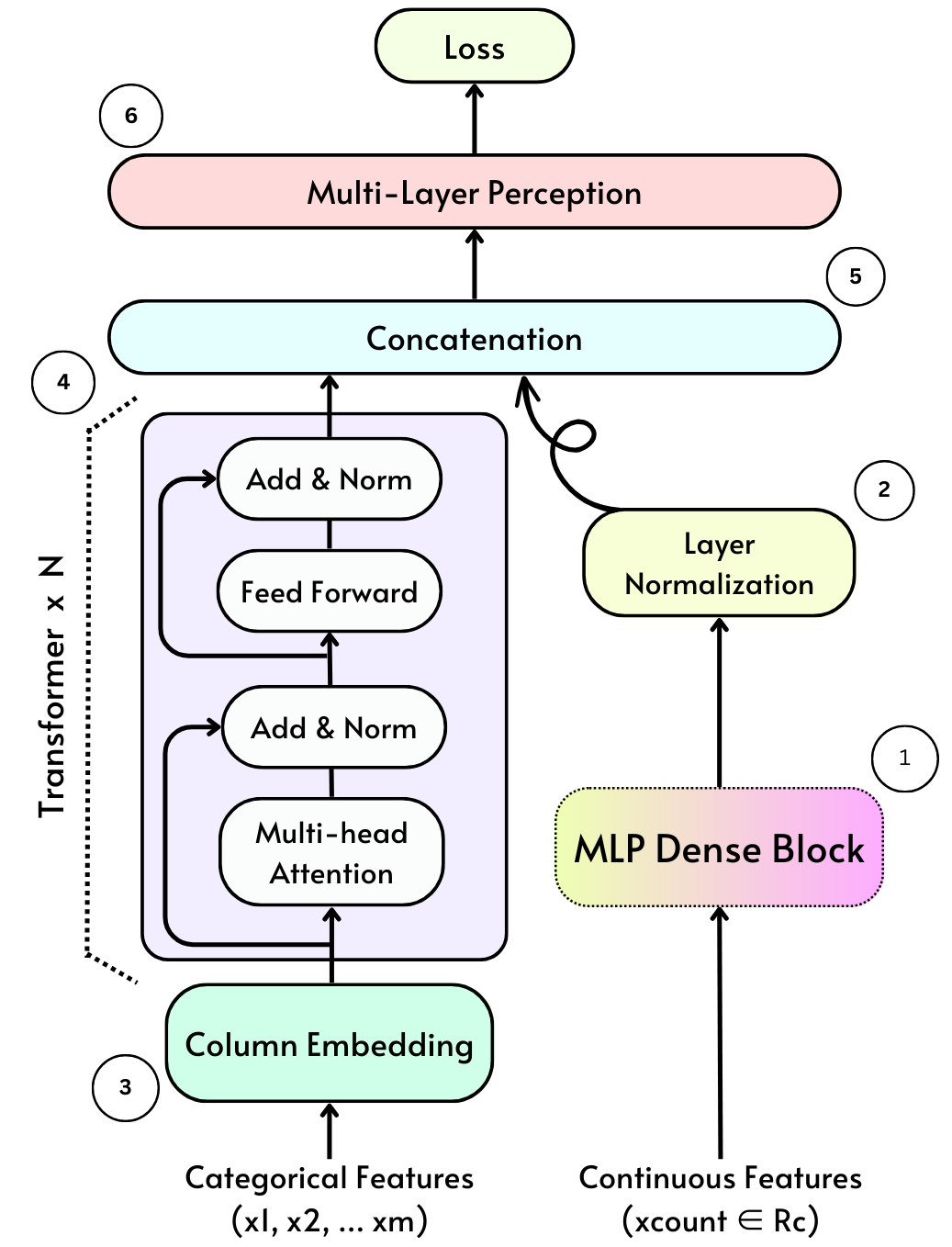}
        \caption{Vanilla MLP TabTransformer Architecture}
        \label{fig:Vanilla MLP TabTransformer Architecture}
    \end{figure}
\hfill\\\\
Let's breakdown the architecture of VM-TabTransformer into six sections:
\hfill\\\\
1. \textbf{MLP Dense Block}: The first step in the VM-TabTransformer is converting the numerical or continuous values from scalar representation to a dense vector contextualization, which has been achieved by using MLP.
\hfill\\\\
2. \textbf{Dense Numerical or Continuous Feature Normalization}: The second step in the VM-TabTransformer architecture involves normalizing the dense numerical features. This normalization process ensures that the dense numerical features have a consistent scale and range which can improve the learning process.
\hfill\\\\
3. \textbf{Categorical Feature Embedding}: The categorical features are converted from scalar to dense vectors representations which are often stated as embeddings. Embeddings capture the semantic meaning and relationships between different categories. This process helps in converting the categorical features into a continuous representation which can be processed by the Transformer.
\hfill\\\\
4. \textbf{Transformer Blocks for Contextual Embeddings}: The embedded categorical features are then passed through multiple Transformer blocks, typically denoted as N as shown in figure.Transformer blocks consist of self attention layers with a feed forward NN. Self-attention mechanisms helps model to capture dependencies and relationships between different categorical features, generating contextual embeddings. The Transformer blocks are stacked to capture increasingly complex relationships and dependencies.
\hfill\\\\
5. \textbf{Concatenation of Contextual Categorical and Dense Numerical Features}: After obtaining the contextual embeddings from the Transformer blocks, they are concatenated or combined with the normalized dense numerical features. This concatenation combines the improved contextual information from the categorical embeddings with the numerical features, which allows the model to utilize both types of knowledge for making predictions.
\hfill\\\\
6. \textbf{MLP for Prediction}: The concatenated features are then passed to a Multi-layer Perceptron architecture. Where the MLP contains of multiple layers of interconnected neurons that can learn complex relationships and patterns in the concatenated feature representation. MLP layer processes the concatenated features to generate the final predictions for the desired task like classification or regression.
\hfill\\\\
Hence to summarize, by incorporating contextual embeddings and dense numerical vectors, the MLP can predict outcomes with greater accuracy. The Vanilla-MLP-TT thus combines Transformers and MLP to convert standard categorical embeddings into contextual ones and a separate MLP helps to convert the numerical values to dense vectors. This fusion enables the TabTransformer to capture dependencies and contextual information, leading to improved performance when learning from categorical and dense numerical features and making more precise predictions.

\subsection{MLP-based - TabTransformer (MLP-TT)}
Here MLP-based refers to the conversion process performed on numerical by a MLP model for converting the scalar numerical values to dense vectors and then concatenating them with the categorial features which are later passed to a Transformer.
\hfill\\\\
The MLP-TT combines Transformers and Multi-layer Perceptron (MLP) to improve the handling of categorical and numerical features. In the Binned-TT, the numerical values were converted to categorical values which losses the core values of the continuous values. Hence to enhance the Binned-TT, the MLP-based-TabTransformer is introduced which process the numerical values using a MLP dense layer for converting the scalar values to dense vectors and those values are concatenated with the categorical values which are passed for the contextual embedding. 
\hfill\\\\
In this process, the MLP-TT model’s Transformer assigns weights to categories based on their relationships with one another, which are then applied to the categorical and dense continuous embeddings. The transformation of embeddings incorporates an attention-based weighting mechanism, allowing the model to capture relevant contextual information. These transformed embeddings of categorical and continuous values provide enhanced contextual information for both the features. 
\hfill\\\\
Hence to sum-up, the MLP-based TabTransformer fuses Transformers and MLP to convert standard categorical and numerical embeddings into contextual values which are used by the Transformer.
\hfill\\\\
Now, let’s understand the architecture of MLP-based-TabTransformer, architecture can be seen in Figure 3.8:
\hfill\\\\
Let's breakdown the architecture of MLP-bases-TabTransformer into six sections:
\hfill\\\\
1. \textbf{MLP Dense Block}: The first step in the MLP-TabTransformer is converting the numerical or continuous values from scalar representation to a dense vector contextualization, which has been achieved by using MLP.
\hfill\\\\
2. \textbf{Dense Numerical Feature Reshaping}: The second step in the MLP-TabTransformer architecture involves reshaping the dense numerical features. This reshaping process ensures that the dense numerical features have a consistent scale and range which are capable of concatenating with the categorical features.
\hfill\\\\
3. \textbf{Concatenation of Categorical and Dense Numerical Features}: After obtaining the dense numerical values from the reshape block, they are concatenated or combined with the categorical features. This concatenation combines the improved continuous or numerical values and categorical values, which allows the model to utilize both types of features for creating contextual embeddings.
\hfill\\\\
4. \textbf{Feature Embedding}: The concatenated features are converted from scalar to dense vectors representations which are often stated as embeddings. Embeddings capture the semantic meaning and relationships between different categories. This process helps in converting the categorical features into a continuous representation which can be processed by the Transformer.
\hfill\\\\
5. \textbf{Transformer Blocks for Contextual Embeddings}: The embedded features are then passed through multiple Transformer blocks, typically denoted as N as shown in figure. Each Transformer block consists self-attention layers and feed-forward NN. Self-attention mechanism allows model to capture dependencies and relationships between different features, generating contextual embeddings. The Transformer blocks are stacked to capture increasingly complex relationships and dependencies.
\hfill\\\\
    \begin{figure}[h!]
        \centering
        \includegraphics[width=0.65\linewidth]{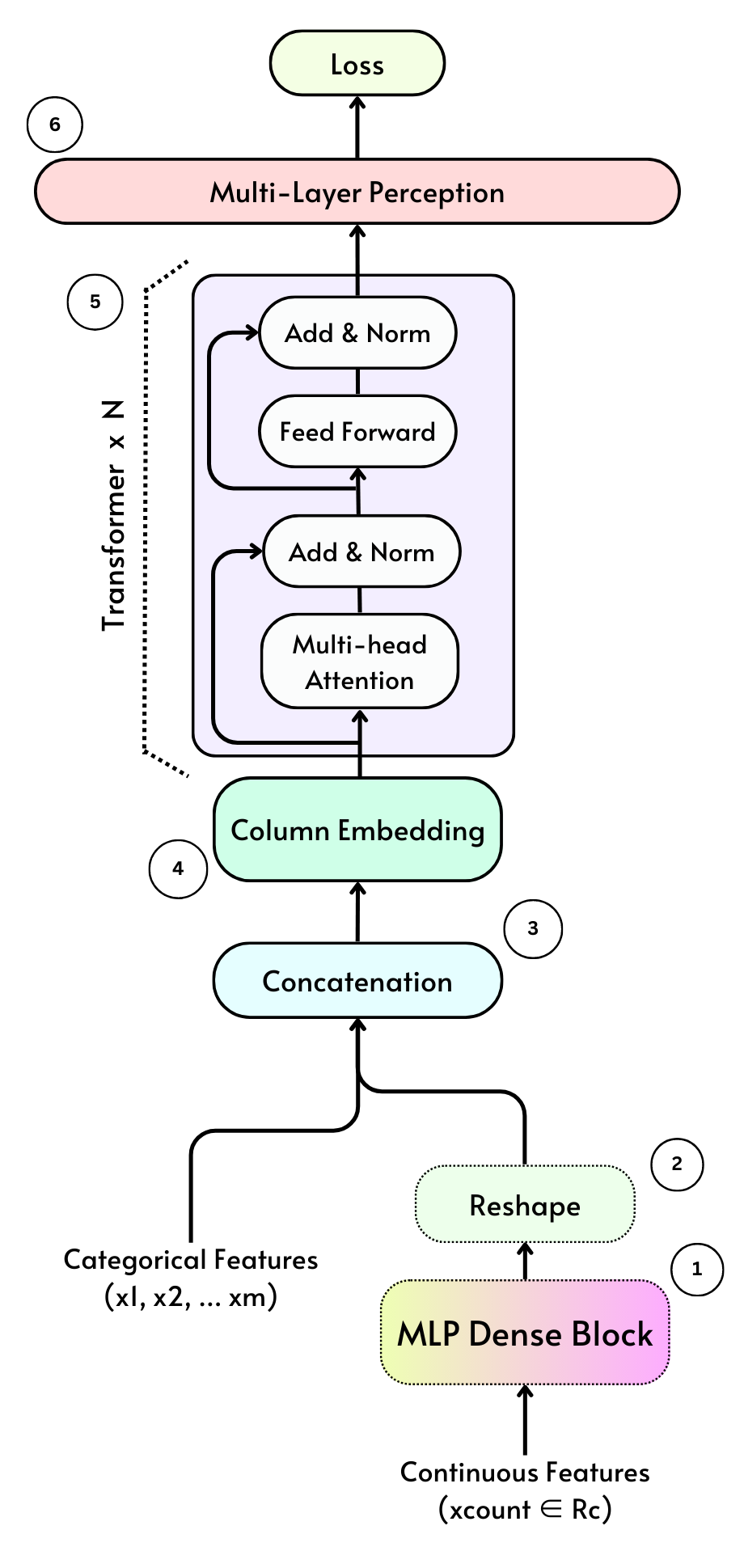}
        \caption{MLP Based TabTransformer Architecture}
        \label{fig:MLP Based TabTransformer Architecture}
    \end{figure}
\hfill\\\\
6. \textbf{MLP for Prediction}: The embedded features are then passed to a Multi-layer Perceptron architecture. Where the MLP contians of multiple layers of interconnected neurons that can learn complex relationships and patterns in the concatenated feature representation. MLP layer processes the features to generate the final predictions for the desired task like classification or regression.
\hfill\\\\
Hence to summarize, by incorporating categorial and dense numerical vectors embedding, the MLP can predict outcomes with greater accuracy. The MLP-TT thus combines Transformers and MLP to convert standard categorical and numerical embeddings into contextual ones. This fusion enables the TabTransformer to capture dependencies and contextual information, leading to improved performance when learning from categorical and dense numerical features and making more precise predictions.
\hfill\\\\
So, we have so far four different variants of the TabTransformer to experiment with. Now we will proceed on how Self-supervised learning works on a TabTransformer.

\section{Self-supervised Learning - TabTransformer}

In this section of paper, we will understand on how Vanilla-TabTransformer has been trained with self-supervised learning approach. The Self-supervised learning approach will remain the same for all of the other variants of the TabTransformer.
\hfill\\\\
In the training process of Vanilla-TabTransformer with a self-supervised learning approach two main tasks are performed: Self-supervised pre-training and Supervised fine-tuning. Before diving into self-supervised learning, the dataset needs to be divided based on the chosen approach.
\hfill\\\\
\textbf{Approach 1 - Random Splitting}: In this approach, the dataset is randomly split into three portions: 60\% and 40\%. The 60\% of the portion is used for self-supervised training, where the model learns to capture meaningful patterns and representations without any labeled information. 
\hfill\\\\
The remaining 40\% portion is further divided into two splits: 10\% and 30\%. The 10\% split is used for supervised fine-tuning, where the model is trained on labeled data for a specific downstream task. The 30\% split is used for evaluating the performance of the trained model on unseen data.
\hfill\\\\
\textbf{Approach 2 - Domain-Based Splitting}: In this approach, the dataset is segregated based on domains. For example, using the Adult Income dataset, the gender column can be chosen as the domain. The dataset is split into two domains: male and female. The male domain is considered the source domain and is used for self-supervised training. The model learns from the source domain to capture representations and patterns. 
\hfill\\\\
The female domain is considered the target domain. It is further divided into two splits: 10\% and 90\%. The 10\% split is used for supervised fine-tuning, where the model is trained on labeled data specific to the target domain. The 90\% split is used for evaluating the model's performance on the target domain.
\hfill\\\\
The reason for having two different approaches is that, in some cases the datasets won’t be having any domain class to segregate hence we will choose the approach based on the type of dataset we are using.
\hfill\\\\
Once the datasets are ready, we have to perform data preprocessing, where we have to apply necessary pre-processing steps, such as normalization, encoding categorical variables, or handling missing values based on the dataset. Further details on the dataset and pre-processing have been mentioned in the experiment section of the paper. Both approaches can be seen below:
    \begin{figure}[h!]
        \centering
        \includegraphics[width=0.95\linewidth]{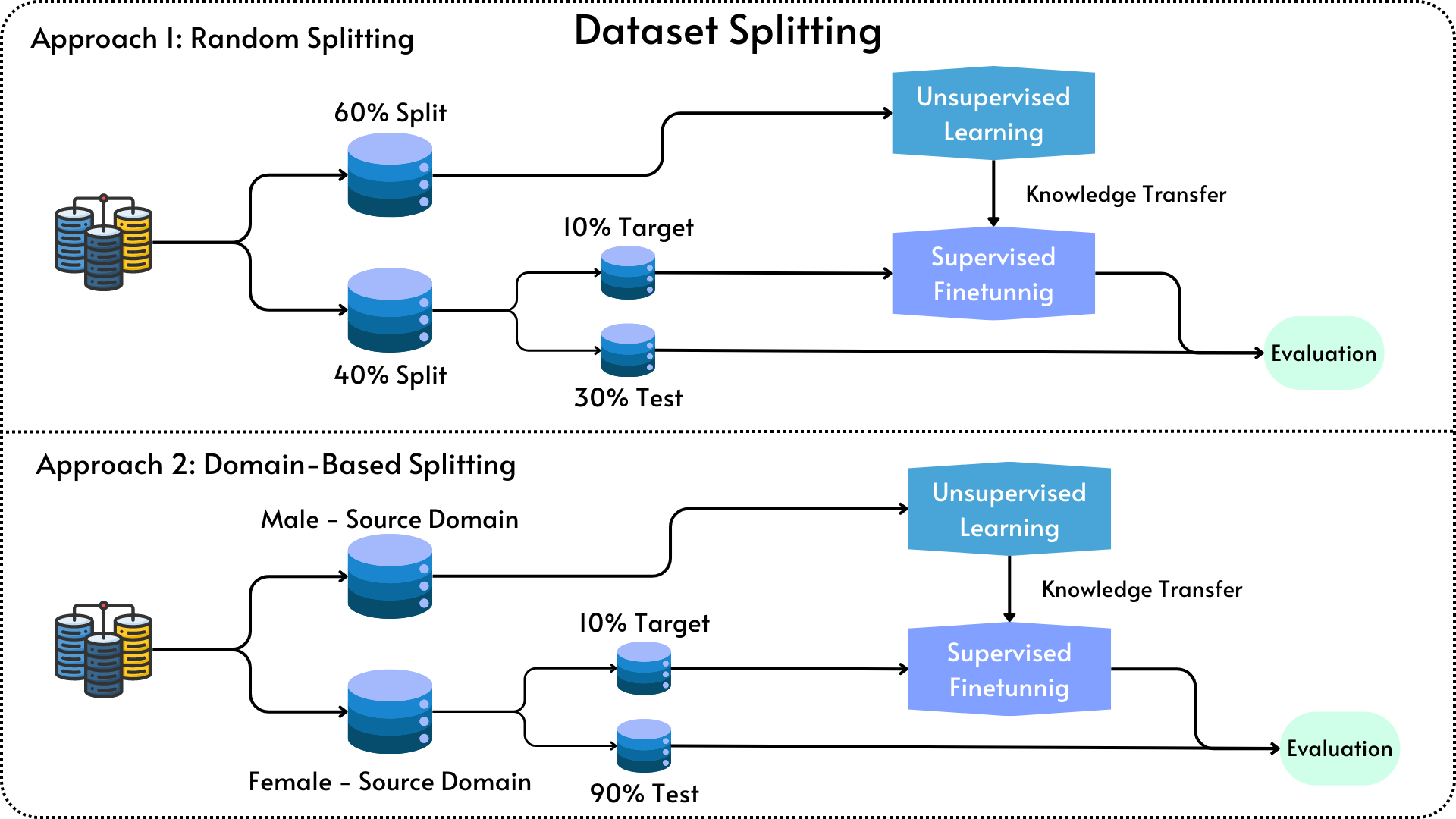}
        \caption{Dataset Splitting}
        \label{fig:Dataset Splitting}
    \end{figure}
\hfill\\\\
Now let’s explore the Self-supervised learning task.
\subsection{Task 1 – Unsupervised Learning} In this task, the TabTransformer model is trained in an unsupervised manner. The goal is to capture meaningful patterns and representations from the input data without using any labeled information. There are various steps which are required to perform the training they are as follows:
\hfill\\\\
\textbf{Step 1}: Defining the TabTransformer architecture, for our experiments we have used the architecture defined in the original paper by the authors.
\hfill\\\\
\textbf{Step 2}: Declaring the model training configuration setups, for our experiments we have used various configurations which are as follows, more details are mentioned in the appendix.
\hfill\\\\
The training settings, which include number of epochs, batch-size, and steps per epoch based on volume of training data, are defined. Later the model is then constructed using the chosen optimizer, learning rate, and weight decay. For our experiments, a CosineDecay schedule is used to set the learning rate, which decreases over time. AdamW, an improved variant of the Adam optimizer that takes weight decay into account, is the optimizer that is used. Mean Squared Error is used for error loss calculation as we are having the target values as scalarized values.
\hfill\\\\
Finally, the model is built with the designated optimizer, loss function, and additional metrics for the evaluation, such as Root Mean Squared Error and Mean Absolute Error.
\hfill\\\\
\textbf{Step 3}: Masking the data and target generation, Masking is one of the applications which is heavily been used in BERT pre-training often called as Masked Language Modelling or MLM. For our task we have also used the MLM approach for performing the training on the TabTransformer. 
\hfill\\\\
In the context of the TabTransformer masking refers to the process of randomly hiding or replacing certain elements of the input data during training. This technique is commonly used in Masked Language Modeling (MLM) tasks in the pre-training phase of BERT.
\hfill\\\\
In the Vanilla-TabTransformer, we apply a similar approach to perform self-supervised learning. During the training process, a portion of the input data is randomly masked and replaced with a special mask token. The original values of the masked positions are then treated as target values. The objective is to train the model to predict the original values based on the contextual information provided by the other features.
\hfill\\\\
By incorporating this masking approach, the TabTransformer learns to capture the dependencies and relationships between the features in the input data. It allows the model to understand the contextual information and fill in the missing values or predict the masked positions accurately.
\hfill\\\\
Hence, masking in the TabTransformer involves randomly hiding or replacing elements of the input data and training model to predict the original values based on the masked inputs. This SSL approach help's our model to learn contextual information and improves its ability to handle missing values or make predictions on incomplete dataset. 
\hfill\\\\
For our experiments, we have masked 20\% of the trained dataset as masked tokens. For example, during the training process a portion of the input data is randomly masked by replacing the original values with a special mask token. For example, if we have a feature with the values [5, 3, 8, 2, 6], the masked version might look like [5, 3, [MASK], 2, 6]. The original values of the masked positions are kept as target values for the model to predict. The masking process can be seen below:
\hfill\\\\
\textbf{Step 4}: Training Objective for TabTransformer, here model aims to learn how to reconstruct original values of the masked features using the context set by the other features during training. We enable the model to capture the dependencies and relationships between the features in the input data by exposing it to masked data and training it to predict the masked values.
\hfill\\\\
\textbf{Step 5}: Loss-Function, we must define suitable loss function in order to evaluate the model's performance for the self-supervised task. MSE loss is one frequently used loss function for masked approaches. Hence, the difference between the masked features predicted values and their original values is measured by this loss function.
    \begin{figure}[h!]
        \centering
        \includegraphics[width=0.68\linewidth]{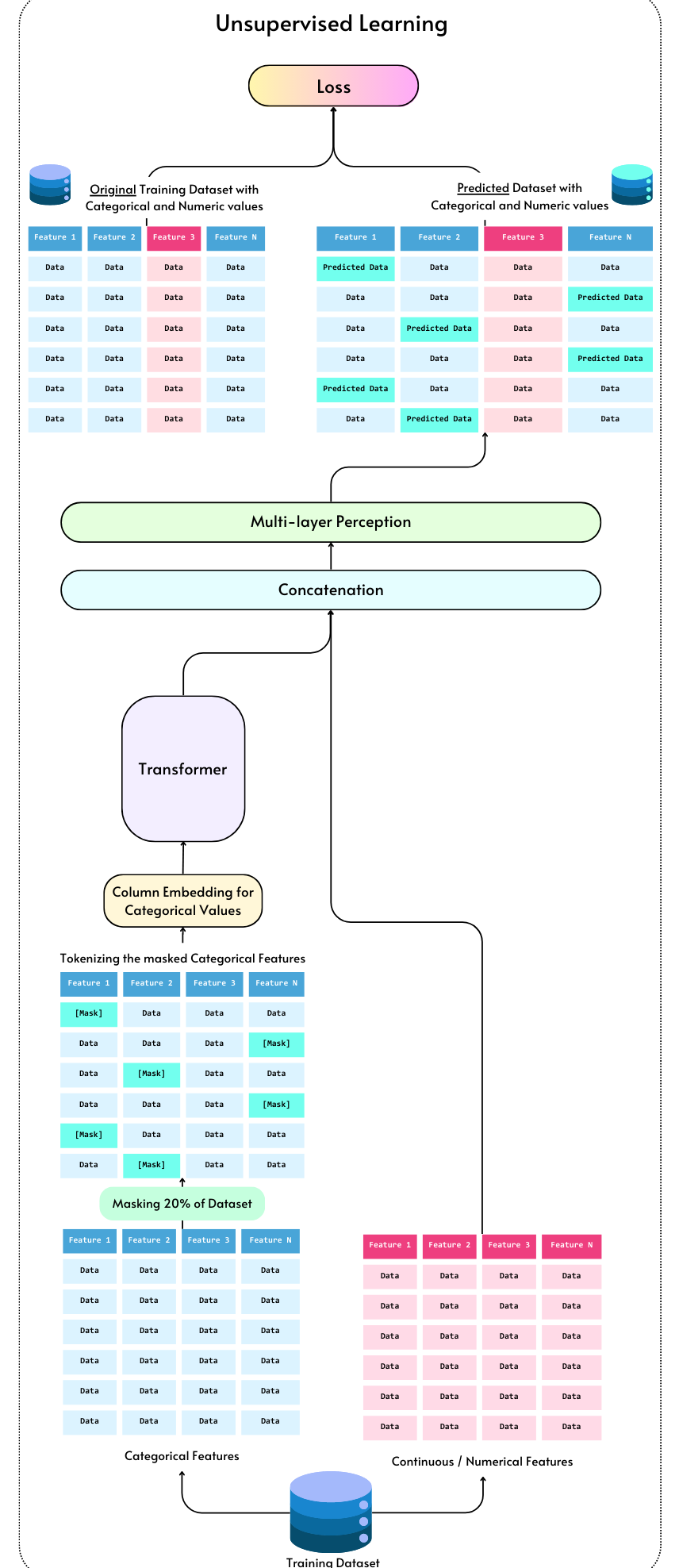}
        \caption{Unsupervised Learning}
        \label{fig:Unsupervised Learning}
    \end{figure}
\hfill\\\\
The model adjusts its parameters during training in an effort to reduce this loss and boost prediction accuracy. The TabTransformer learns to accurately predict the masked positions or fill in the missing values based on the context provided by the other features by optimizing the loss function.
\hfill\\\\
The TabTransformer model can effectively learn contextual information, handle missing values, and make precise predictions on incomplete datasets by incorporating this masking approach and defining an appropriate loss function. The Self-supervised pre-training task can be seen below in Figure 3.9. Whereas, here the masked data is compared to original data during training process. Model is trained to predict the original values of the masked tokens. This is done by using a loss-function which helps to measure difference between predicted and original values. The model is updated to minimize the loss function.
\hfill\\\\
Once, the Vanilla-TT is been trained in an unsupervised fashion with minimal loss. We can execute the process called as saving weights from the trained model and use the same model weights to perform the supervised fine-tuning task based on our target domain.
\hfill\\\\
To elaborate, after training the Vanilla-TabTransformer model in an unsupervised manner using the masking task, we can save the weights of the trained model. These saved weights capture the learned representations and contextual information from the self-supervised training. Subsequently, we can utilize these saved weights and perform supervised fine-tuning on a target domain. Fine-tuning refers to the process of further training pre-trained model on specific task or dataset will adapt it to the target domain. In the context of the TabTransformer, fine-tuning involves using the saved weights from the self-supervised training and updating the model parameters based on the labeled data in the target domain.
\subsection{Task 2 - Supervised Finetuning}
Here's an overview of the process:
\textbf{Step 1 - Saving the Weights}:
After training the Vanilla-TabTransformer model using self-supervised learning approachs, save weights of trained model. This allows you to store the learned representations and configurations of the model.
\hfill\\\\
\textbf{Step 2 - Target-Domain Data}:
Preparing the labeled-data for the target domain which we want to perform supervised fine-tuning on. This data should have the input features along with their corresponding target label.
\hfill\\\\
\textbf{Step 3 - Model Architecture}:
Setting up the TabTransformer model architecture with the same configuration as the previously trained Vanilla-TabTransformer model. Load the saved weights into the model to initialize its parameters.
\hfill\\\\
\textbf{Step 4 - Loss Function}:
Define a suitable loss function for the specific task in the target domain. This could be-cross-entropy loss for classification tasks or mean squared error loss for regression tasks.
\hfill\\\\
\textbf{Step 5 - Fine-tuning}:
Train the TabTransformer model using labeled data in target domain. The model parameters are initialized with the saved weights, and during fine-tuning, they are further updated based on the labeled data and the chosen loss function. Using an optimization algorithm, we have used AdamW, to update the model parameters and minimize the loss function. Fine-tuning allows the model to adapt and specialize its learned representations to the specific task and data in the target domain.
\hfill\\\\
    \begin{figure}[h!]
        \centering
        \includegraphics[width=0.80\linewidth]{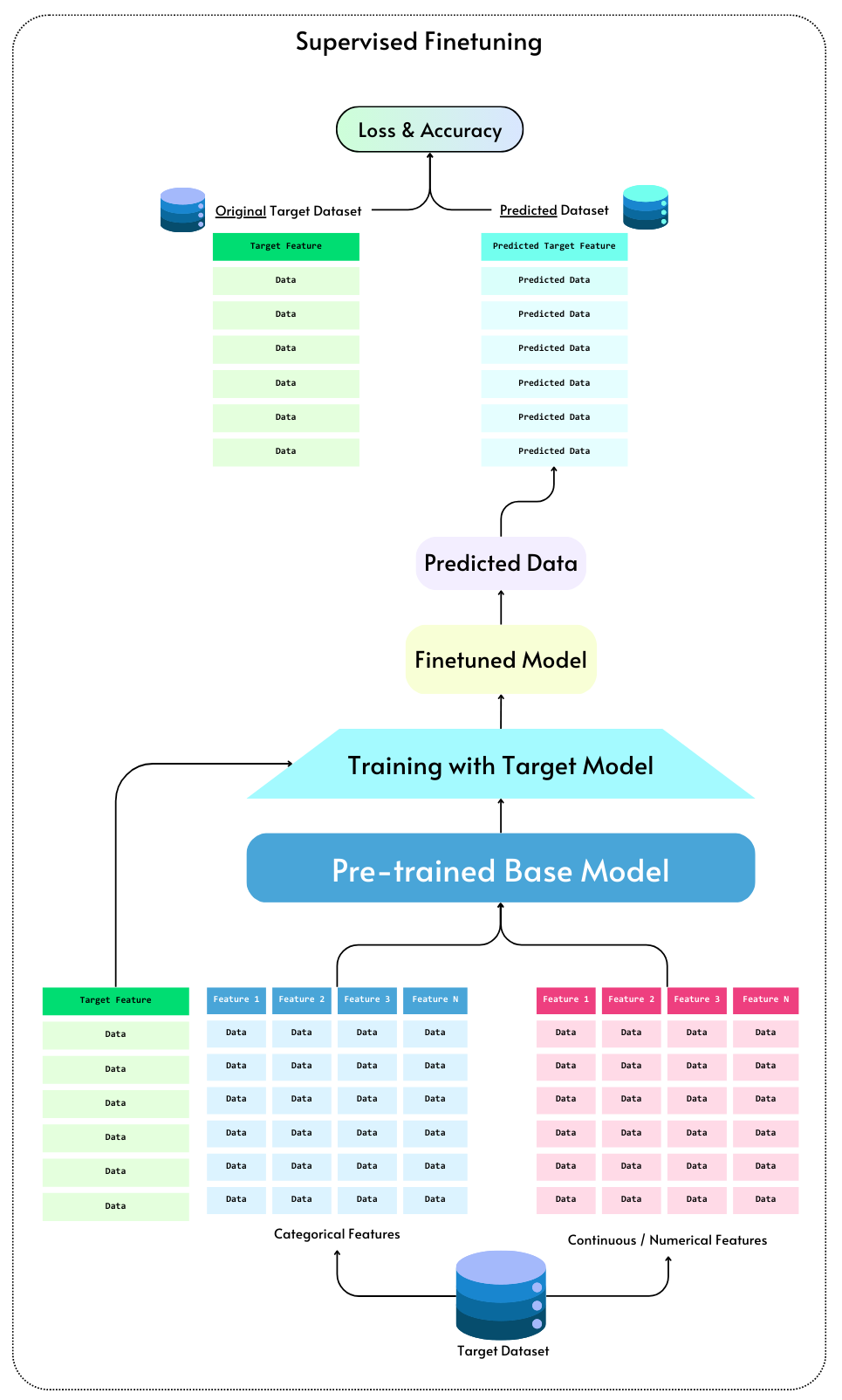}
        \caption{Supervised Finetuning}
        \label{fig:Supervised Finetuning}
    \end{figure}
\hfill\\\\
\textbf{Step 6 - Evaluation}: 
After fine-tuning, we have evaluated the performance of model on separate test dataset in the target domain as mentioned in the dataset approaches. And the evaluation metrics are based on the task to assess the model's performance and generalization ability.
\hfill\\\\
Hence, Vanilla-TabTransformer can efficiently learn useful representations from the input data and optimize them for a particular downstream task by combining self-supervised pre-training and supervised fine-tuning. So, when self-supervised learning is used with real-world datasets this approach can enhance the model's functionality and capacity for generalization for each task. Both the task for Self-supervised Learning can be seen below:
\hfill\\\\
    \begin{figure}[h!]
        \centering
        \includegraphics[width=0.90\linewidth]{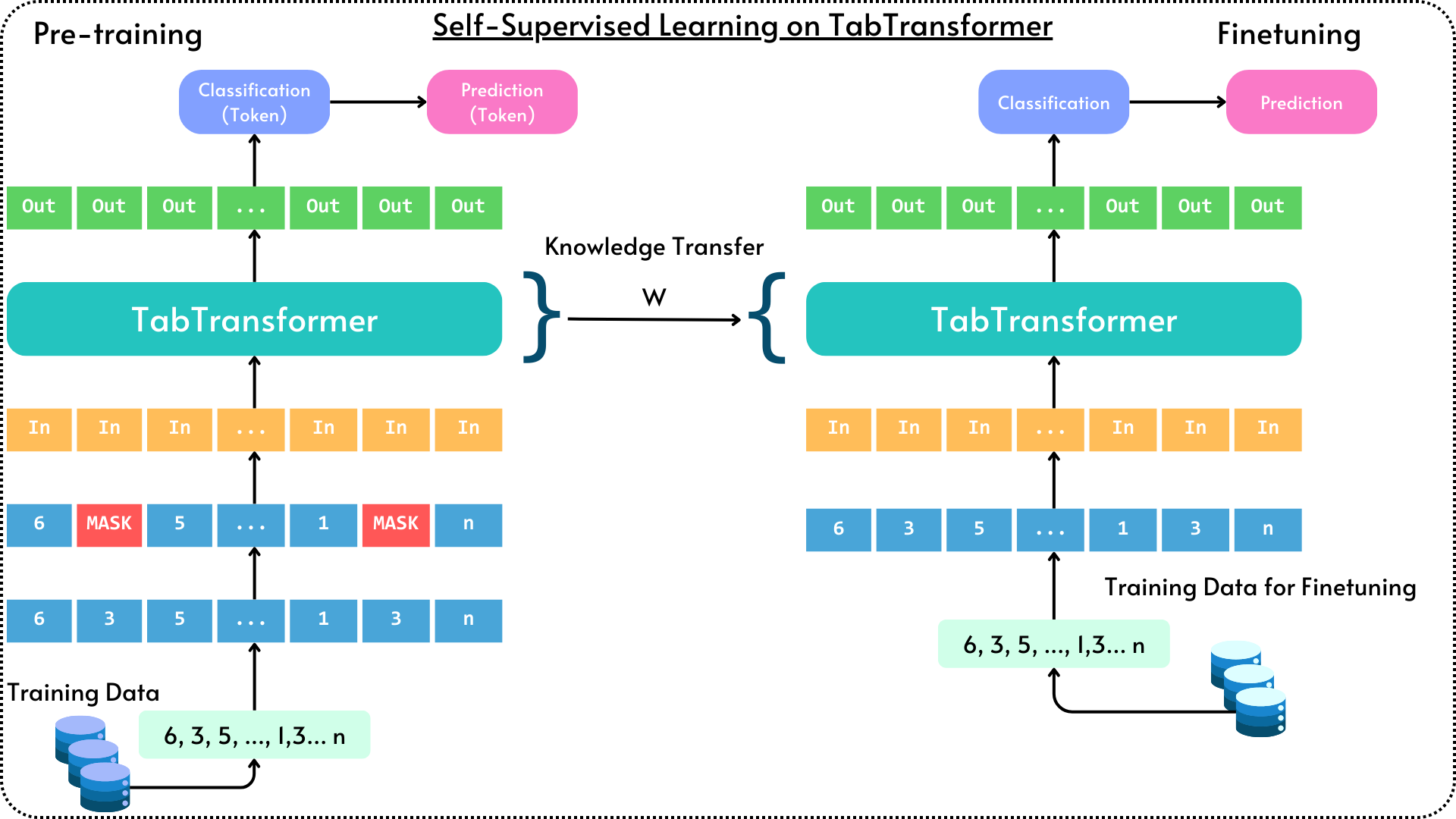}
        \caption{Self-supervised Learning on TabTransformer.}
        \label{fig:Self-supervised Learning on TabTransformer.}
    \end{figure}
\hfill\\\\
    \begin{figure}[h!]
        \centering
        \includegraphics[width=0.95\linewidth]{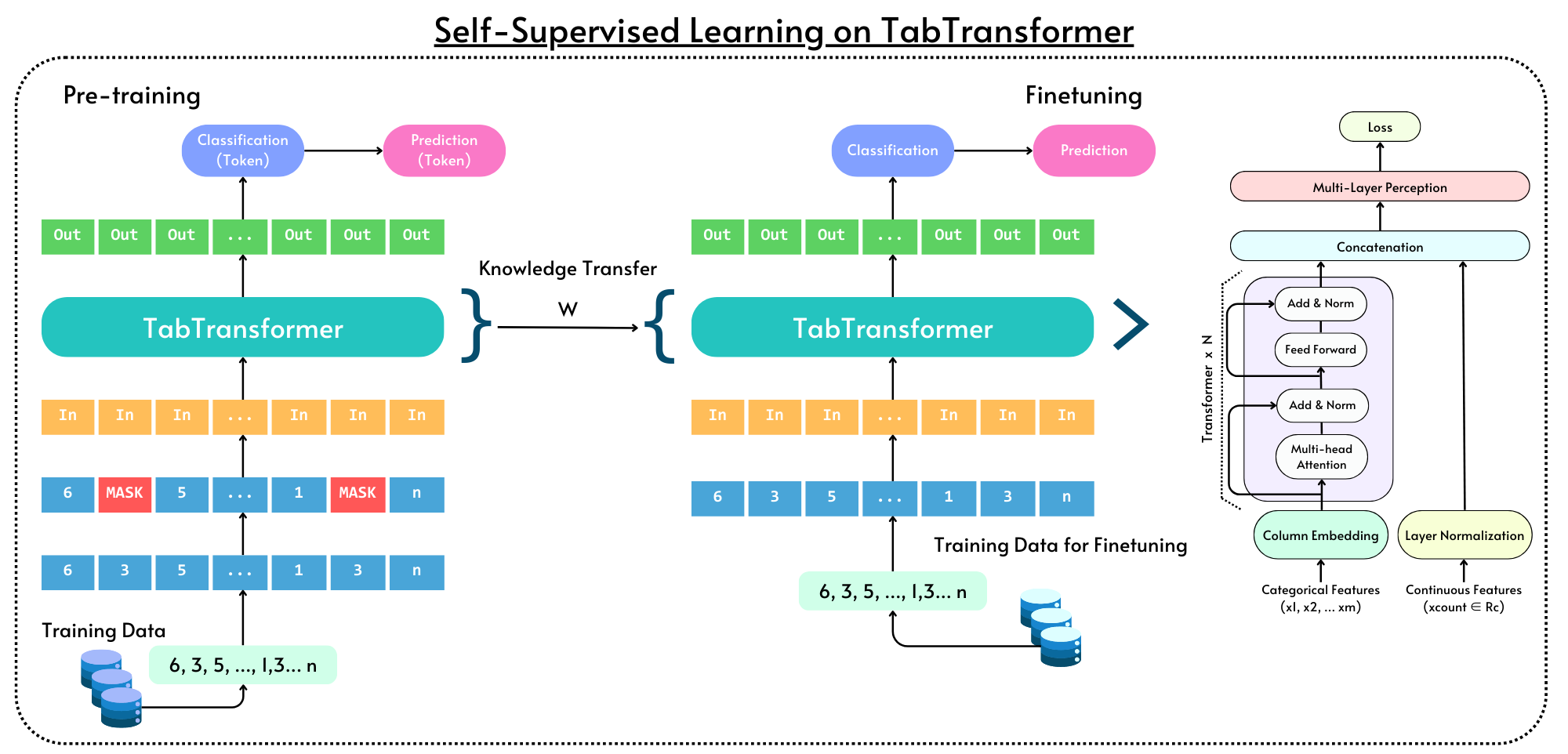}
        \caption{Self-supervised Learning on TabTransformer}
        \label{fig:Self-supervised Learning on TabTransformer}
    \end{figure}

\chapter{Experiments \& Results}
\label{chap4}

\section{Datasets}

For our experiments I have chosen three datasets, which are as follows: 
\subsection{1. Adult Census Income:} 
One of the most common dataset used in machine learning and data analysis tasks is the Adult Census Income Dataset, also referred as the "Census Income" or "Census" dataset. The dataset includes data on people that were obtained from a variety of sources, majorly from the U.S. Census Bureau database from 1994. The dataset is often used in classification tasks, where the objective is to determine whether an individual's income exceeds a predetermined threshold (typically \$50,000 per year) based on a set of features(\cite{adult_data}).
\hfill\\\\
Here are some details about the Adult Census Income Dataset:
\hfill\\\\
\textbf{01. Size}: The dataset consists of approximately 32561 records or rows, each representing an individual set of data.
\hfill\\\\
\textbf{02.	Attributes}: Each instance in the dataset is described by 15 attributes or features, as follows:
\hfill\\\\
•	Age: Age of an individual which is a continuous value.
\hfill\\\\
•	Workclass: Type of work an individual is engaged in.
\hfill\\\\
•	Fnlwgt: Functional data for individuals.
\hfill\\\\
•	Education: Highest level of education completed by an individual.
\hfill\\\\
•	Education-num: Numerical representation of education which is a continuous value.
\hfill\\\\
•	Marital status: The marital status of an individual.
\hfill\\\\
•	Occupation: Occupation of the individual.
\hfill\\\\
•	Relationship: Relationship status of an individual
\hfill\\\\
•	Race: Race of the individual.
\hfill\\\\
•	Sex, Capital-gain and loss and country are all continuous values.
\hfill\\\\
•	Income: Income level of an individual, classified as ">50K" (income exceeds \$50,000) or "<=50K" (income is at most \$50,000).
\hfill\\\\
\textbf{03. Categorical Attributes}: The adult dataset consist of 10 categorical features, as shown below:
\hfill\\\\
    \begin{figure}[h!]
        \centering
        \includegraphics[width=0.95\linewidth]{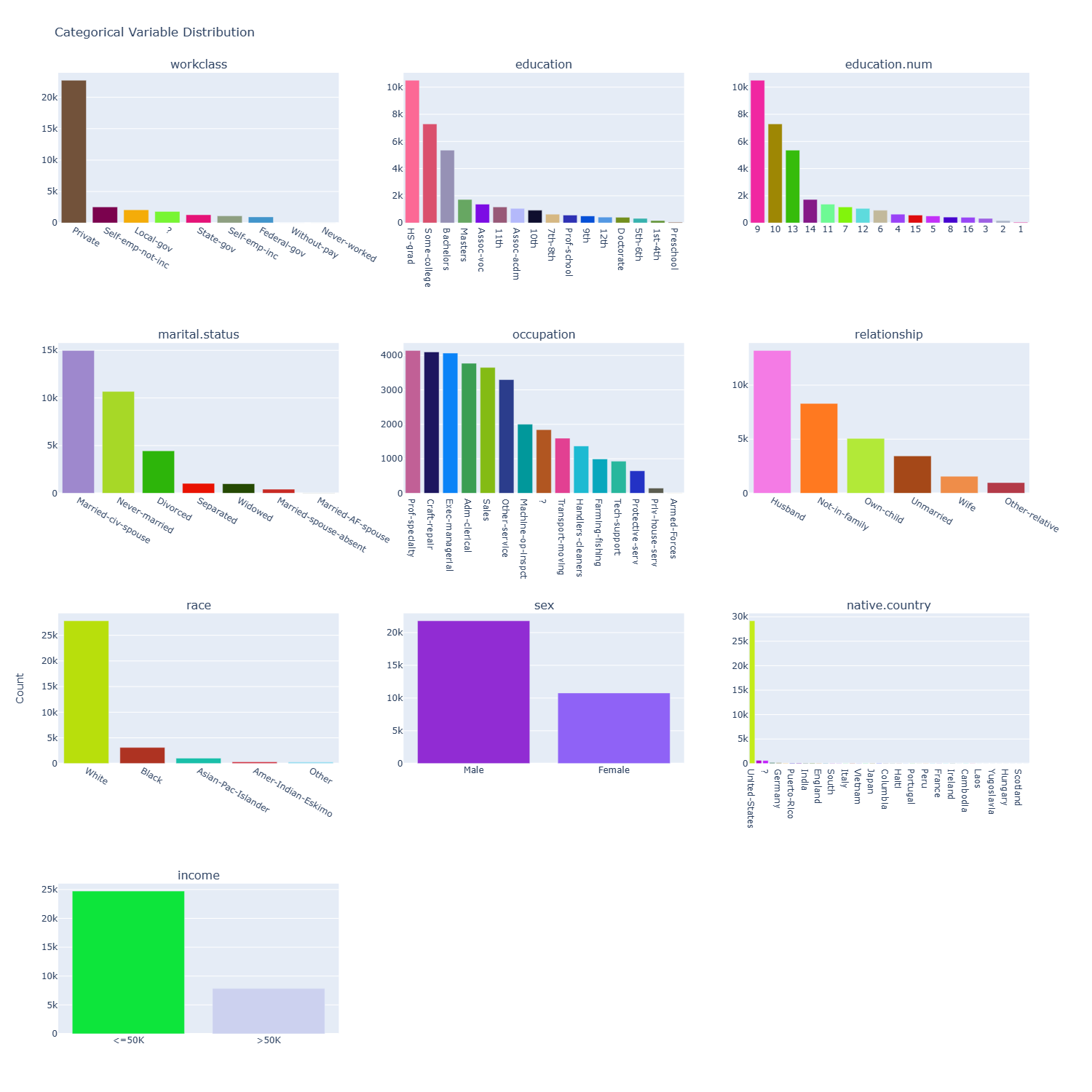}
        \caption{Adult income dataset: Categorical values}
        \label{fig:Adult income dataset: Categorical values}
    \end{figure}
\hfill\\\\
\textbf{04. Numerical / Continuous Attributes}: The adult dataset consist of 5 categorical features, as shown below:
\hfill\\\\
\textbf{05. Target Feature}: Here, for our experiments the target label for supervised fine-tuning is "income" which can be considered as a classification-based feature.
\hfill\\\\
    \begin{figure}[h!]
        \centering
        \includegraphics[width=0.95\linewidth]{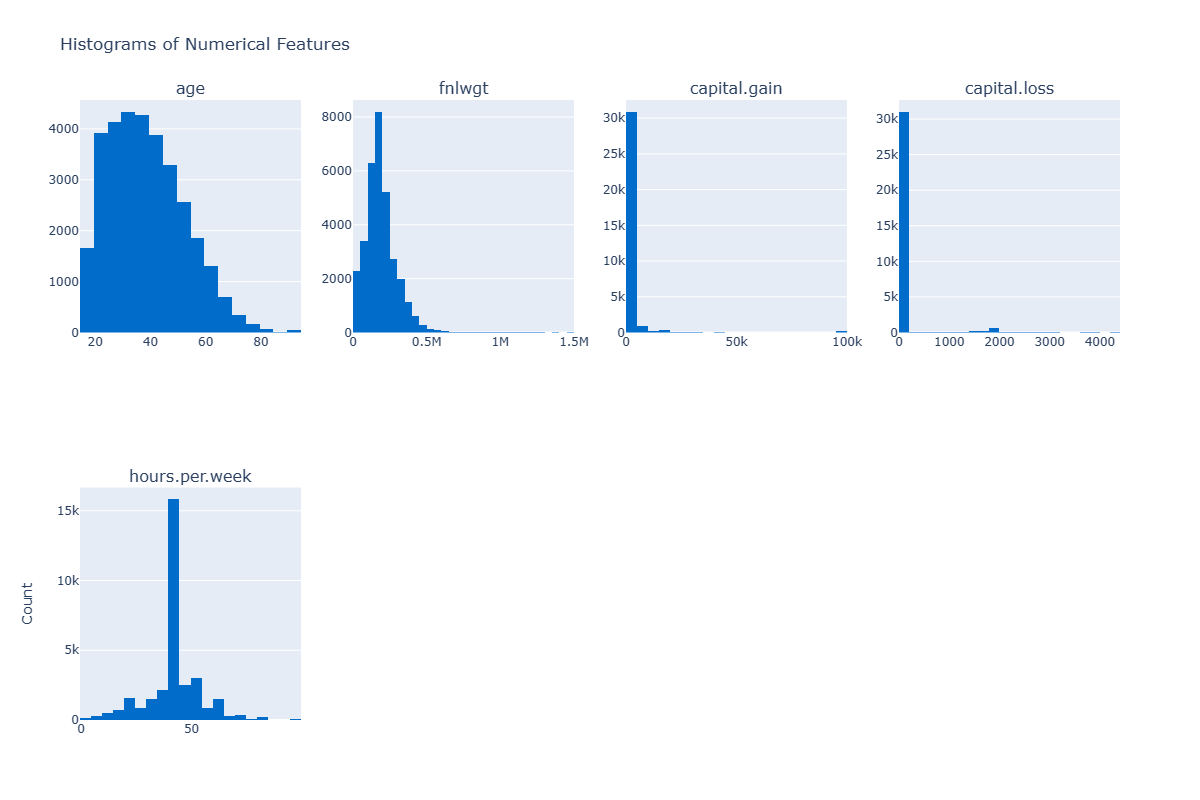}
        \caption{Adult income dataset: Continuous values}
        \label{fig:Adult income dataset: Continuous values}
    \end{figure}
\hfill\\\\
    \begin{figure}[h!]
        \centering
        \includegraphics[width=0.70\linewidth]{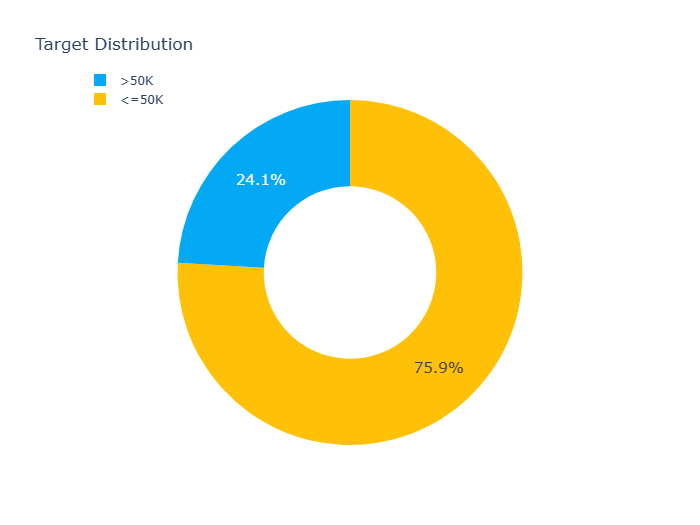}
        \caption{Adult income dataset: Target feature for finetuning}
        \label{fig:Adult income dataset:  Target feature for finetuning}
    \end{figure}
\hfill\\\\

\clearpage

\subsection{2. California Housing Prices:} 

In data analysis, the "California Housing Prices Dataset" is one of the frequently used dataset. Which offers details on real estate costs across California's various regions. This dataset falls under regression problem based dataset, which aims to predict the median house value based on other attributes(\cite{california_1997}).
\hfill\\\\
Here are some details about the California Housing Prices Dataset:
\hfill\\\\
\textbf{01. Size}: The dataset consists of approximately 20639  records or rows, each representing an individual set of data.
\hfill\\\\
\textbf{02.	Attributes}: Each instance in the dataset is described by 10 attributes or features, the major features are as follows:
\hfill\\\\
•	median\_income: Median income.
\hfill\\\\
•	housing\_median\_age: Median age of house.
\hfill\\\\
•	total\_rooms: Average number of rooms.
\hfill\\\\
•	total\_bedrooms: Average number of bedrooms.
\hfill\\\\
•	population: Total population in the block.
\hfill\\\\
•	latitude: Latitude coordinate of the block.
\hfill\\\\
•	longitude: Longitude coordinate of the block.
\hfill\\\\
•	median\_house\_value: house value or price.
\hfill\\\\
•	ocean\_proximity: proximity of house near the ocean (continuous value).
\hfill\\\\
\textbf{03. Categorical Attributes}: The California house dataset consist of 1 categorical feature, as shown below:
    \begin{figure}[h!]
        \centering
        \includegraphics[width=0.45\linewidth]{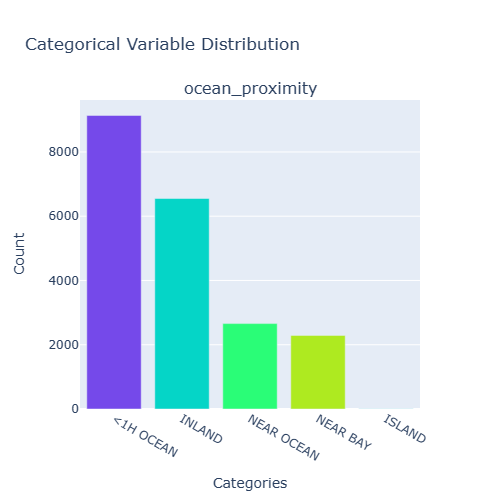}
        \caption{California Housing dataset: Categorical values}
        \label{fig:California Housing dataset: Categorical values}
    \end{figure}
\hfill\\\\
\textbf{04. Numerical / Continuous Attributes}: The California house dataset consist of 9 numerical features, as shown below:
\hfill\\\\
    \begin{figure}[h!]
        \centering
        \includegraphics[width=0.90\linewidth]{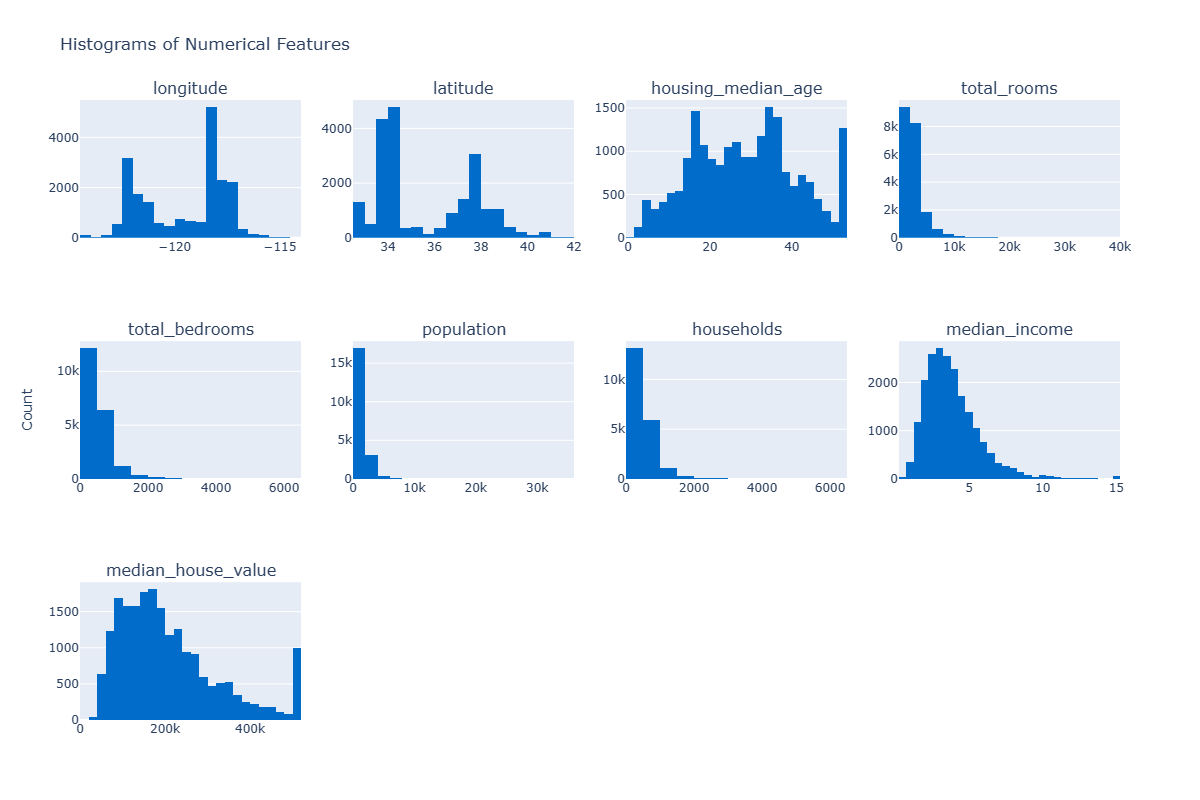}
        \caption{California Housing dataset: Continuous values}
        \label{fig:California Housing dataset: Continuous values}
    \end{figure}
\hfill\\\\
\textbf{05. Target Feature}: Here, for our experiments the target label for supervised fine-tuning is "median\_house\_value " which can be considered as a regression-based feature.
\hfill\\\\
    \begin{figure}[h!]
        \centering
        \includegraphics[width=0.70\linewidth]{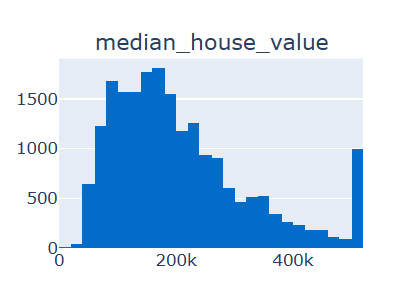}
        \caption{California Housing dataset: Target feature for finetuning}
        \label{fig:California Housing dataset:  Target feature for finetuning}
    \end{figure}

\subsection{3. Breast Cancer Wisconsin (Diagnostic):} 
"The Breast Cancer Wisconsin (Diagnostic) Dataset, Dr. William H. Wolberg from University of Wisconsin Hospitals sorted it"(\cite{cancer_2011}). The dataset includes data on cell nuclei that was taken from pictures of breast tumour samples. The objective is to categorise tumours as benign (non-cancerous) or malignant (cancerous). 569 instances or samples make up the dataset, which has 32 attributes or features. The sample's ID numbers appear as the 1st attribute, followed by diagnosis (M for malignant, B for benign), and then other 30 additional attributes that describe various aspects of the cell nuclei. Size, shape, and texture measurements are some of these characteristics(\cite{cancer_2011}). 
\hfill\\\\
Here are some details about the Breast Cancer Dataset:
\hfill\\\\
\textbf{01. Size}: The dataset consists of approximately 569  records or rows, each representing an individual set of data.
\hfill\\\\
\textbf{02.	Attributes}: Each instance in the dataset is described by 32 attributes or features, the major features are as follows:
\hfill\\\\
•	The 1st column is ID number of sample.
\hfill\\\\
•	The 2nd column is diagnosis (M = malignant, B = benign), which is target variable to predict, will be using for supervised finetuning.
\hfill\\\\
•	The remaining 30 attributes (3 to 32) are valued feature computed from cells nucleus characteristics. These features as follows, "radius, texture, perimeter, area, smoothness, compactness, concavity, symmetry, fractal dimension", etc.
\hfill\\\\
\textbf{03. Categorical Attributes}: The Breast Cancer dataset consist of 1 categorical feature, as shown below:
    \begin{figure}[h!]
        \centering
        \includegraphics[width=0.55\linewidth]{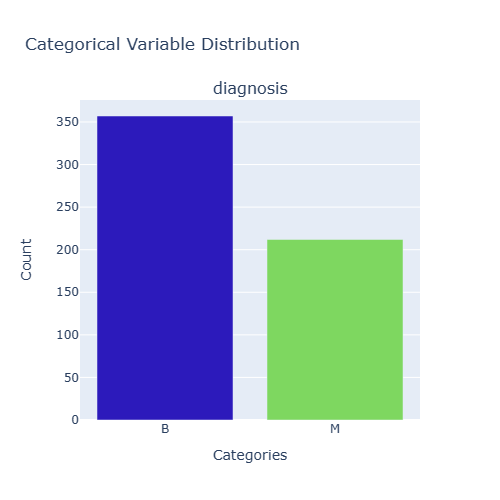}
        \caption{Breast Cancer dataset: Categorical values}
        \label{fig:Breast Cancer dataset: Categorical values}
    \end{figure}
\hfill\\\\
\textbf{04. Numerical / Continuous Attributes}: The Breast Cancer dataset consist of 31 numerical features, as shown below:
    \begin{figure}[h!]
        \centering
        \includegraphics[width=0.90\linewidth]{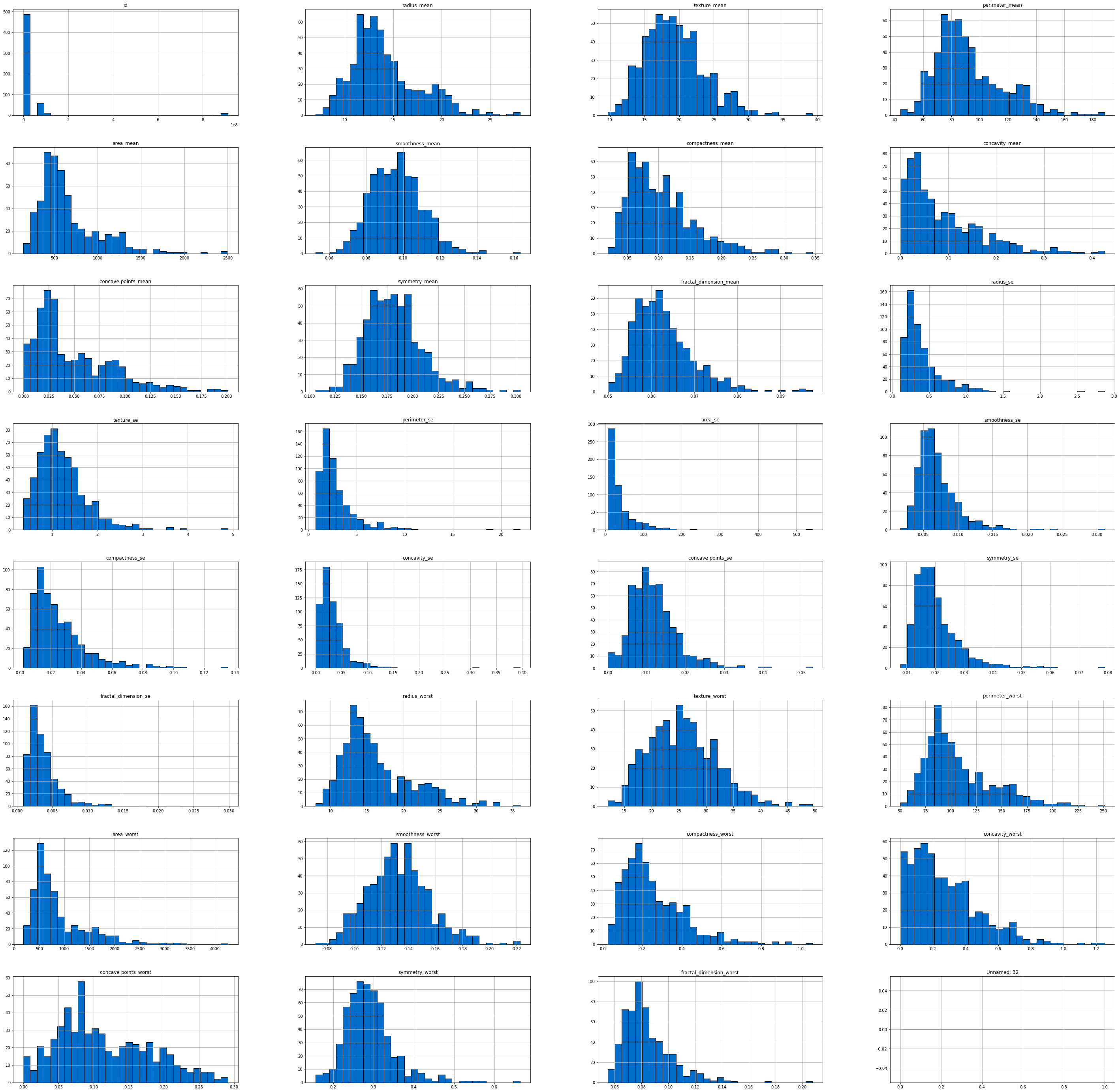}
        \caption{Breast Cancer dataset: Continuous values}
        \label{fig:Breast Cancer dataset: Continuous values}
    \end{figure}
\hfill\\\\
\textbf{05. Target Feature}: Here, for our experiments the target label for supervised fine-tuning is "diagnosis " which can be considered as a classification-based feature.
    \begin{figure}[h!]
        \centering
        \includegraphics[width=0.57\linewidth]{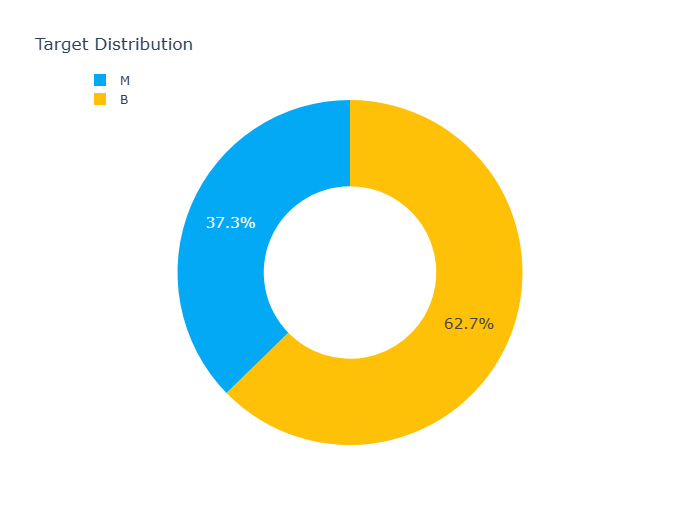}
        \caption{Breast Cancer dataset: Target feature for finetuning}
        \label{fig:Breast Cancer dataset:  Target feature for finetuning}
    \end{figure}
    
\clearpage

\section{Models \& Metrics}
As mentioned in the Methodology section of the paper, we have four variants for the TabTransformer, which were implemented for Self-supervised Learning (SSL) settings.
\hfill\\\\
\textbf{1. Vanilla-TabTransformer: .... SSL (V-TT)}
\hfill\\\\
\textbf{2. Binned-TabTransformer: .... SSL (B-TT)}
\hfill\\\\
\textbf{3. Vanilla-MLP-TabTransformer: .... SSL (VM-TT)}
\hfill\\\\
\textbf{4. MLP-based-TabTransformer: .... SSL (MLP-TT)}
\hfill\\\\
Additionally, for the comparison of the results I have used the following models in Supervised learning (SL) settings, details have been mentioned in appendix:
\hfill\\\\
\textbf{5. Baseline Multi-layer Perception: .... SL (MLP)}
\hfill\\\\
\textbf{6. TabTransformer: .... SL (TT)}
\subsection{Training Methodology}
\textbf{Data Preprocessing.} 
\hfill\\\
Adult dataset, null values were checked and removed from the dataset. Once data was clean, I have splitted the dataset based on Approach 2 - Domain based splitting as mentioned in the Methodology section, which is based on the domains. Here, the domain was gender and based on that the data was bifurcated as per male and female attributes from the source domain. Male source data was used to perform self-supervised pre-training and a fraction of Female data i.e 10\% was used for supervised finetunning, and the remaining data was used for testing the model. Subsequently, label encoding was also performed on the categorical features as it is always easy for a model to work with number's rather than strings and later on MinMaxScaler was performed on the dataset for normalization.  
\hfill\\\\
California Housing Prices \& Breast Cancer  dataset, null values were checked and removed from the dataset. Once data was clean, I have splitted the dataset based on Approach 1 - Random splitting as mentioned in the Methodology section, which is based on the random split of 60\% and 40\%. Here, the whole dataset was bifurcated in two splits, 60\% data was used to perform self-supervised pre-training and a fraction of 40\% data i.e 10\% was used for supervised finetunning, and the remaining data was used for testing the model. The reason for not using the Domain based splitting was that, there were no categorical values to split the dataset based on Approach 2. Subsequently, label encoding was also performed on only 1 remaining categorical feature and later on MinMaxScaler was performed on the dataset for normalization.
\hfill\\\\
\textbf{Loss Function.} 
\hfill\\\
During Self-supervised pre-training, for all three of the datasets the loss function used was Mean Squared Error/L2 Loss. The reason for using the MSE loss was that as we performing the task of predicting the masked values in the dataset and the values were in continuous form or it can be said as a regression based prediction here the self-supervised pre-training is basically a type of unsupervised learning approach where the model can learn the underlying patterns from the predicted dataset.
\hfill\\\\
Whereas, for the supervised finetunning I have used two different losses based on the problem or the target feature we want the model to be tuned on. For example, for the Adult and Cancer dataset the target feature is income or diagnosis which was either greater than 50k or less then 50k or M(malignant) or B(benign) tumor respectively so this can be associated as the binary classification problem. Hence for the binary classification I have used Binary Crossentropy loss. And for the house price prediction I have used MSE loss where we have to predict the house value associated as regression based problem or target.
\hfill\\\\
\textbf{Evaluation Metric.} 
\hfill\\\
For evaluating the Adult and Cancer dataset, I have used Binary Accuracy and Binary Crossentropy loss. Because for the both the datasets we are predicting binary values either 0 or 1. And for the California Housing dataset, I have used MSE and MAE losses as all the models doesn't really predict the exact continuous value for the prediction. So, here the losses will provide estimation on how close the value was predicted for the original value.
\hfill\\\\
\textbf{Cross Validation.} 
\hfill\\\
CV - A frequently used method for assessing performance metrics in tabular data. CV is necessary because there aren't enough data points in datasets of small to medium size to allocate a separate test dataset that isn't used during training. The dataset is randomly divided into five portions of equal size for cross-validation. While the remaining four portions are used to train the model, each portion is alternately used as the validation set. I have implemented 5-fold CV for our problem, the approach can be seen as below:
\hfill\\\\
    \begin{figure}[h!]
        \centering
        \includegraphics[width=0.75\linewidth]{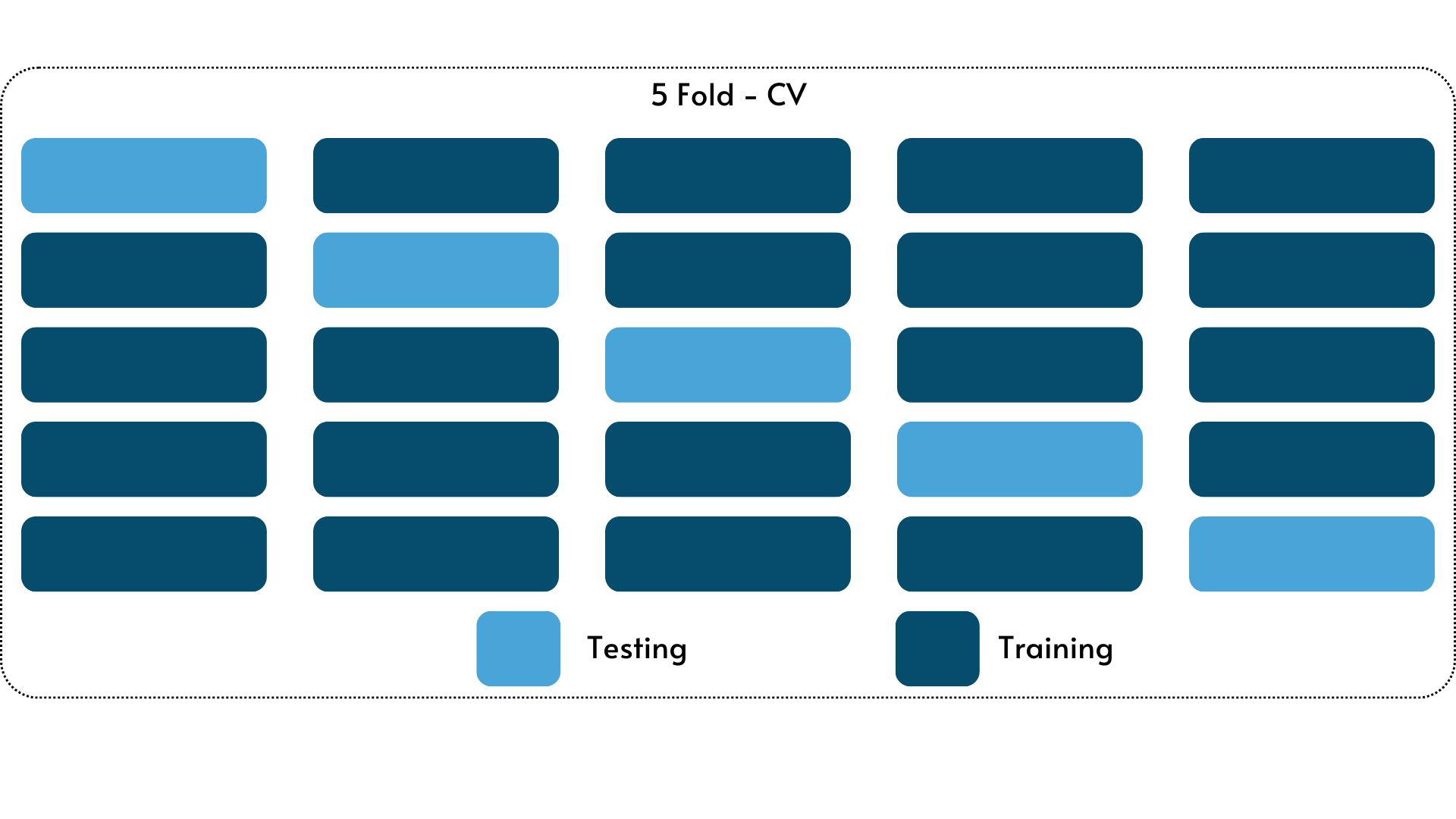}
        \caption{Five - fold Cross Validation}
        \label{fig:Five - fold Cross Validation}
    \end{figure}
\pagebreak
\subsection{Result Analysis}
Hence, based on the aspects aforementioned I have conducted experiments on the three dataset for six different models. The results generated by the experiments is as follows:
\hfill\\\\
    \begin{figure}[h!]
        \centering
        \includegraphics[width=0.95\linewidth]{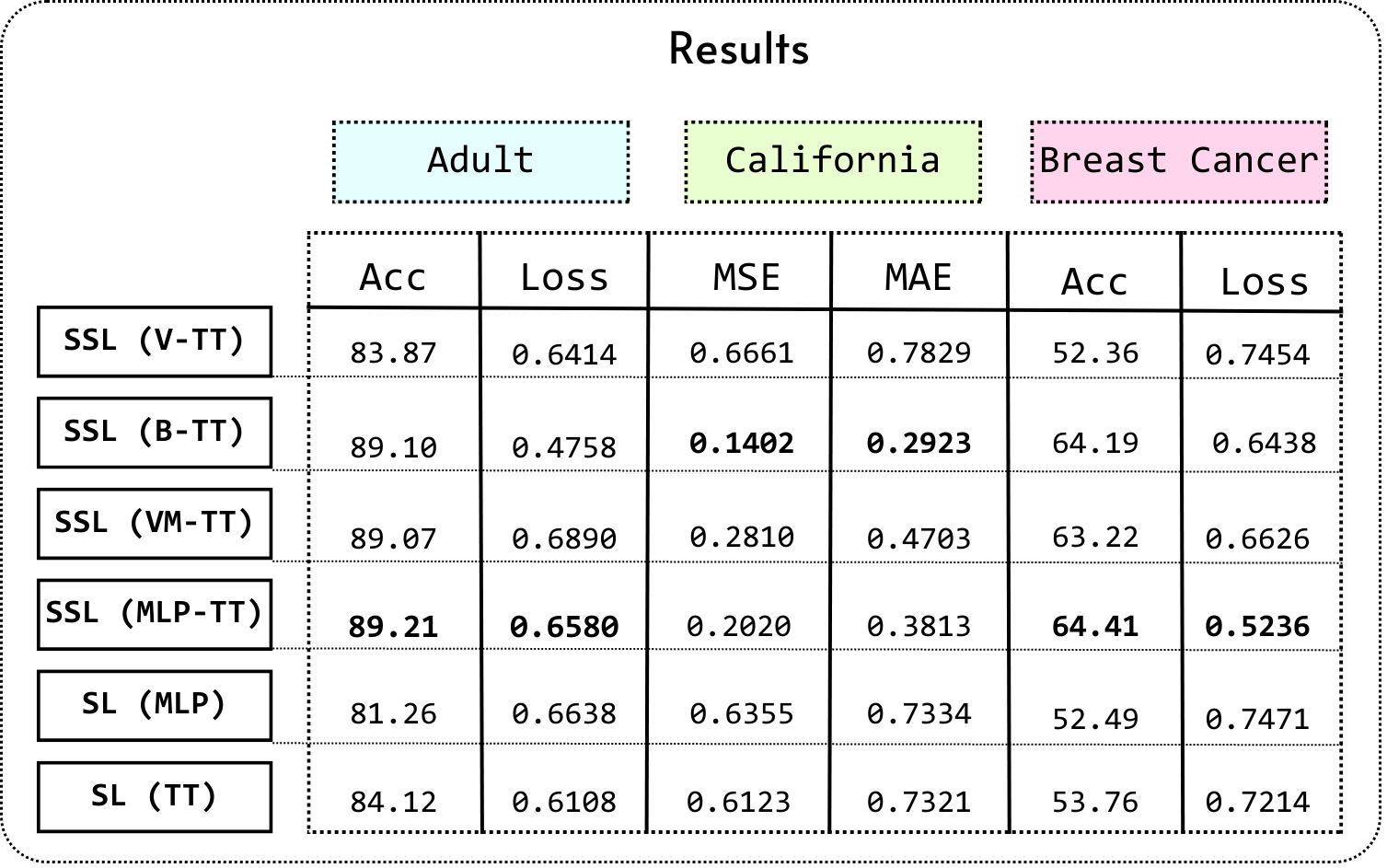}
        \caption{Results}
        \label{fig:Results}
    \end{figure}
\hfill\\\\
The results are been visually plotted for better understanding in the below mentioned image.
    \begin{figure}[h!]
        \centering
        \includegraphics[width=0.85\linewidth]{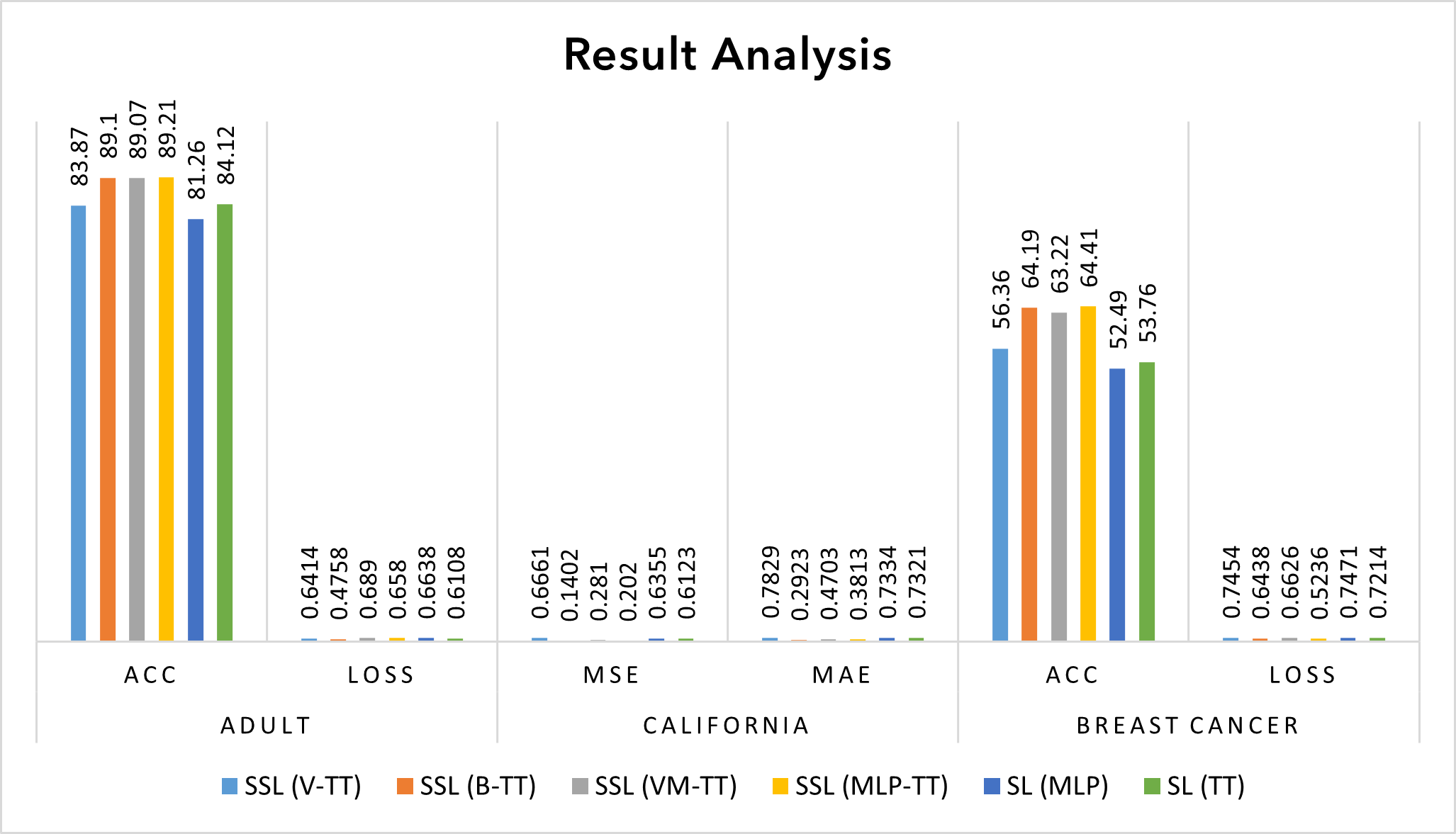}
        \caption{Result Analysis}
        \label{fig:Result Analysis}
    \end{figure}
\hfill\\\\
For our analysis, we will explore each dataset's results.
\hfill\\\\
\textbf{1. Adult Census Income Dataset:} 
\hfill\\\\
    \begin{figure}[h!]
        \centering
        \includegraphics[width=0.95\linewidth]{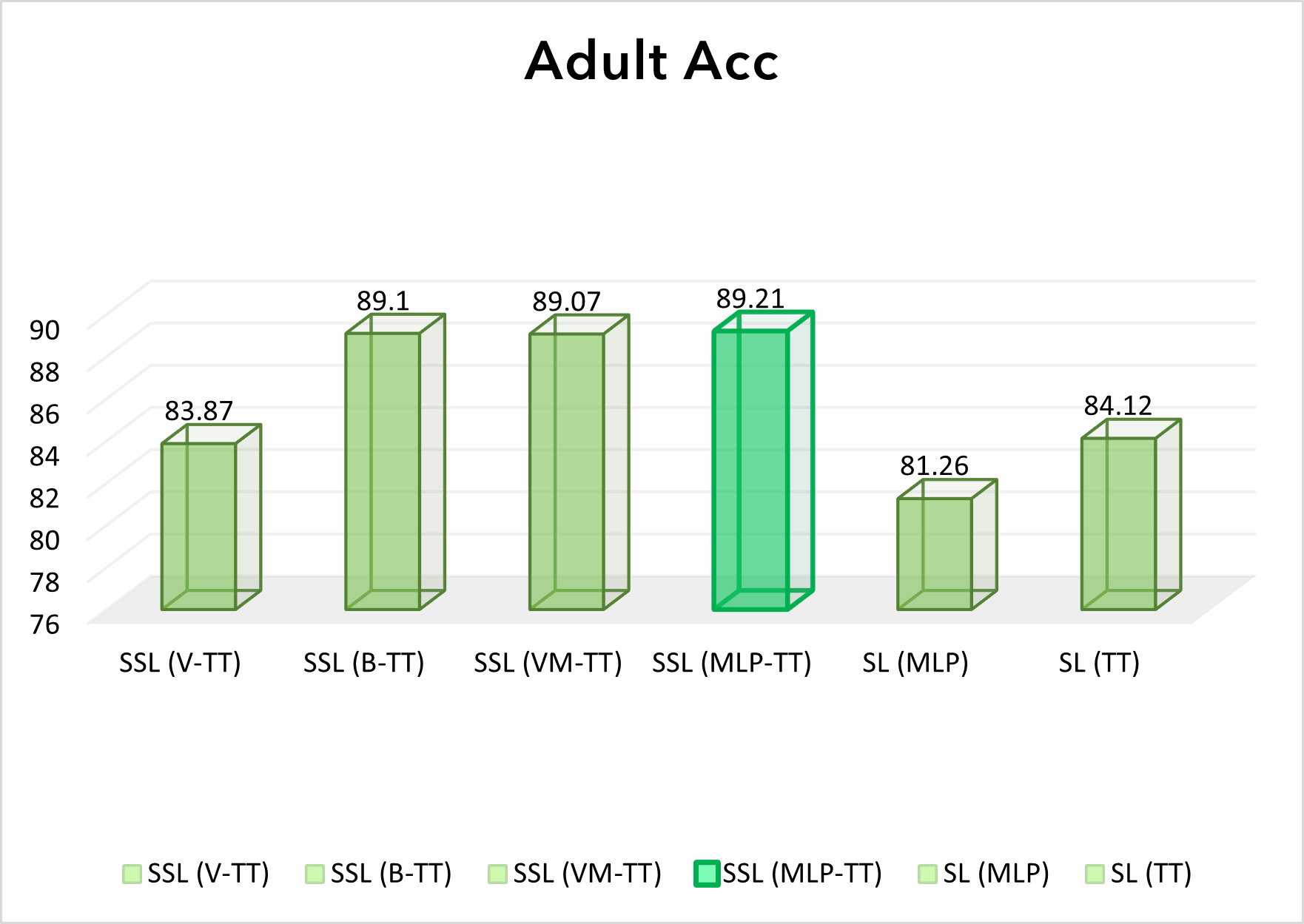}
        \caption{Adult Result Analysis}
        \label{fig:Adult Result Analysis}
    \end{figure}
\hfill\\\\
For the above mentioned Adult dataset it was found that, the SSL (MLP-TT) has performed with an accuracy score of 83.21\% following the loss of 0.6580. It can be stated that the approach for creating the MLP's dense vectors for continuous values and then passing them with the categorical embedding has sucessfuly worked for the model in uderstanding the underlying structure and insights from the dataset. Further training a small fraction of the test dataset was able to finetune the model for the providing a better accuracy as compared to the other TabTransformer variants and supervised base models. Maybe a good reason for such good accuracy on all of the SSL task is the diversity of the dataset. 
\hfill\\\\
The dataset contained of approx. 32k rows with 10 Categorical values and 5 Numerical values, which helped the MLP-TT model's categorical embedding and dense numerical values to out perform supervised base models. Comparing the MLP-TT with the other SSL based model like V-TT, B-TT, VM-TT. The B-TT was second with the accuracy of 89.10\% because of the conversion of the remaining 5 numerical values to categorical values, but converting the values affect the underlying relationship of the dataset and a potential loss in the information. Whereas, the V-TT stated 83.87\% which performed worst than the SL(TT) with 84.12\%, maybe because of the 10\% of the finetuning data where it was not able to accuratly learning the patterns with the small data for tuning as the SL model were trained on 60\% of the dataset as compared to SSL with just 10\% of finetuning.
\hfill\\\\
\hfill\\\\
\textbf{2. California Housing Prices Dataset:} 
\hfill\\\\
    \begin{figure}[h!]
        \centering
        \includegraphics[width=0.95\linewidth]{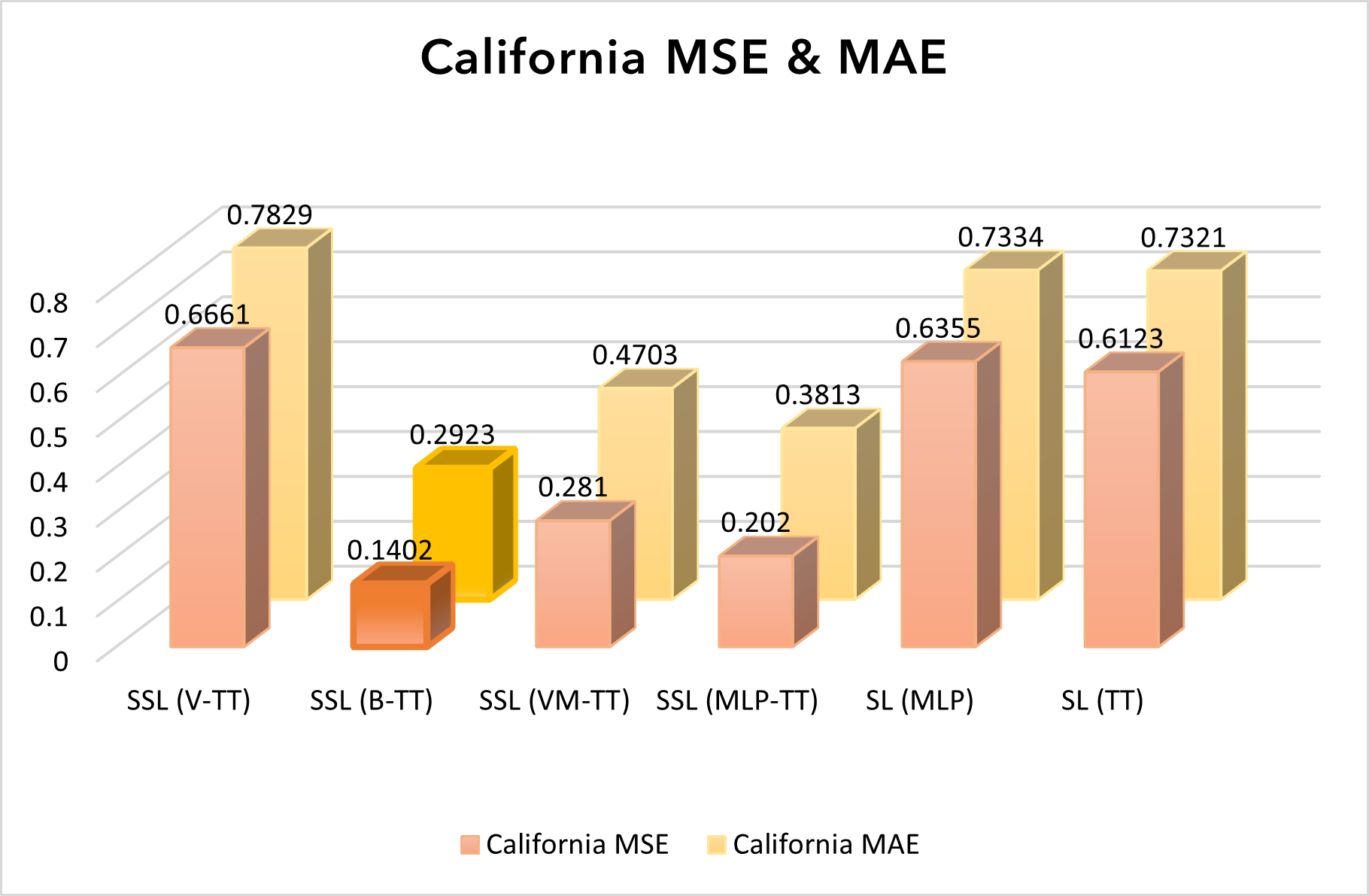}
        \caption{California Housing Result Analysis}
        \label{fig:California Housing Result Analysis}
    \end{figure}
\hfill\\\\
For the above mentioned California dataset it was found that, the SSL (B-TT) has performed with score of 0.1402 MSE and 0.2923 MAE. It can be stated that the approach for creating the bins and converting the numerical values to categorical drastically affects the models performance as the embedding for the categorical values are able to learn useful insights from the transformer. The supervised tuning also performed as aspected for the model learning as it was tuned on a small fraction of the testing dataset which help tuning and minising the loss of the model.
\hfill\\\\
The dataset contained of approx. 20k rows with 1 Categorical value and 9 Numerical values, which helped the B-TT model's categorical embeddings to out perform supervised base models as well. Comparing the B-TT with the other SSL based model like V-TT, MLP-TT, VM-TT. The MLP-TT was second with the score of 0.2020 MSE and 0.3813 MAE because of the conversion of the remaining 9 numerical values to dense values, but converting the values affect the underlying relationship of the dataset and a potential loss in the information. Whereas, the V-TT stated 0.6661 MSE and 0.7829 MAE which performed worst than the SL(TT \& MLP), maybe there were not enough learning from the embedding as it was only having one categorical feature as compared to the Adult dataset. Hence, analysing the result it sheds light on how a diverse dataset like Adult can allow the Transformer embedding can learn the under laying connection between the values.
\hfill\\\\
\hfill\\\\
\hfill\\\\
\textbf{3. Breast Cancer Wisconsin (Diagnostic) Dataset:} 
\hfill\\\\
    \begin{figure}[h!]
        \centering
        \includegraphics[width=0.80\linewidth]{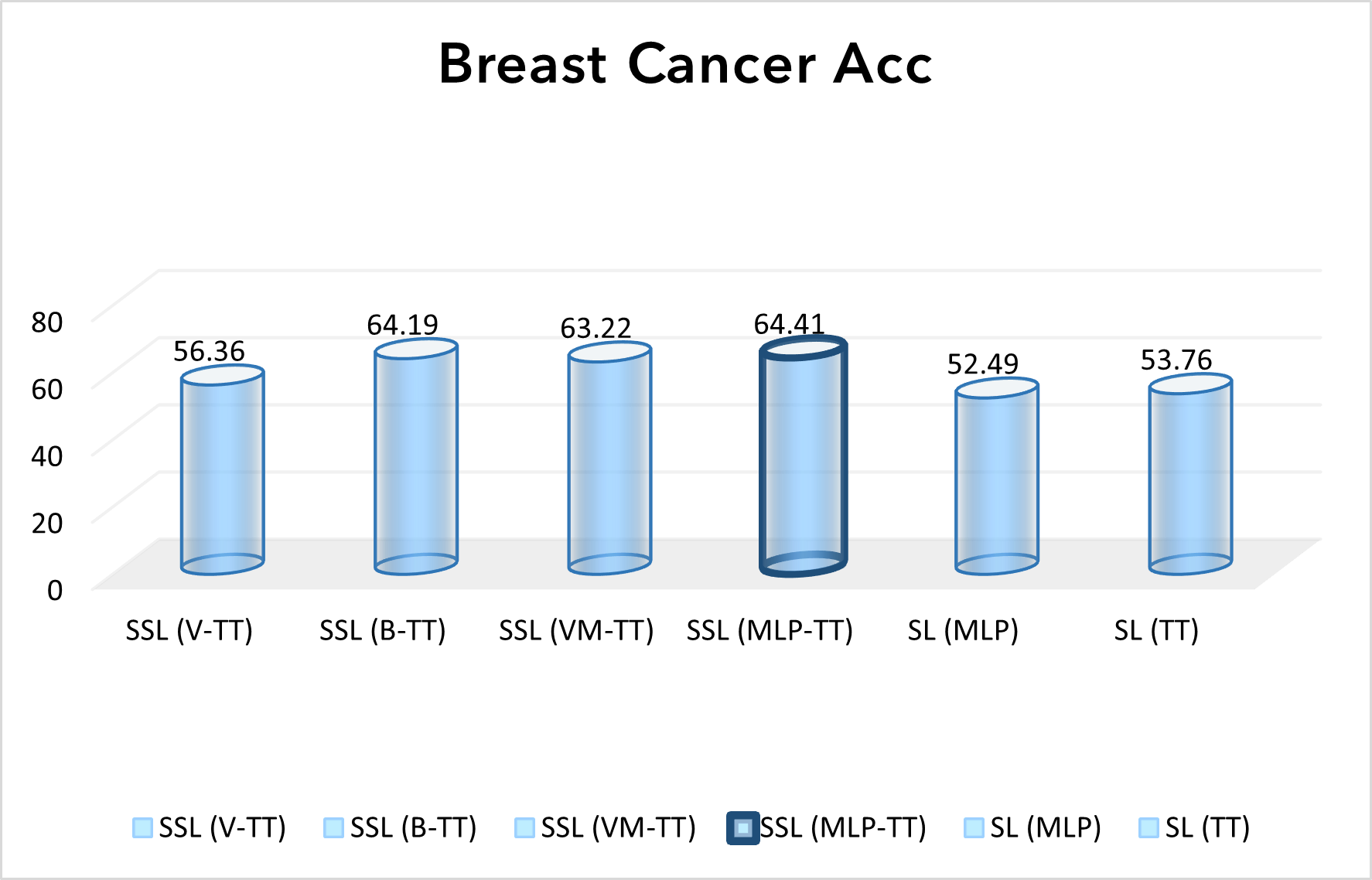}
        \caption{ Breast Cancer Result Analysis}
        \label{fig: Breast Cancer Result Analysis}
    \end{figure}
\hfill\\\\
For the above mentioned Cancer dataset it was found that, the SSL (MLP-TT) has performed with an accuracy score of 64.41\% following the loss of 0.5236. It can be stated that the approach for creating the MLP's dense vectors for continuous values and then passing them with the categorical embedding has sucessfuly worked for the model in uderstanding the underlying structure and insights from the dataset. Further training a small fraction of the test dataset was able to finetune the model for the providing a better accuracy as compared to the other TabTransformer variants and supervised base models. Maybe a good reason for such good accuracy on all of the SSL task is the diversity of the dataset. 
\hfill\\\\
The dataset contained of only 569 rows with 1 Categorical value and 31 Numerical values, which helped the MLP-TT model's categorical embedding and dense numerical values to out perform supervised base models. Comparing the MLP-TT with the other SSL based model like V-TT, B-TT, VM-TT. The B-TT was second with the accuracy of 64.19\% because of the conversion of the remaining 31 numerical values to categorical values, but converting the values affect the underlying relationship of the dataset and a potential loss in the information. The reason for the Cancer dataset having an accuracy of no above 70\% for all of the models is that, all of our model are based on neural networks. Neural networks require a large amount of dataset for learning or training the model. Larger the dataset more the model can learn the underlying structure of the data. Hence, it can be stated as the size of the dataset also plays an important role in performing the tasks.
\hfill\\\\
To summarize the analysis of the results, it can be stated that the dataset's construction and size plays an important role for model like TabTransformer which have seperate input construction for categorical and numerical values.
\hfill\\\\

\chapter{Discussion}
\label{chap5}

\section{Limitations}
There are several limitations, of our aforementioned models and results. As, mentioned in the analysis section that the size and the structure of the datasets play an important role in the field of deep learning, our all models were based on a TabTransformer structure having various setting for numerical and categorical input construction. So, it is important to have a diverse dataset containing a mix and a proportional blend of numerical and categorical values and the dataset should be large enough for the training as the transfer learning plays important role while finetuning the supervised model.
\hfill\\\\
The self-supervised learning methodology used in this study also has a limitation. As it doesn't require labelled data while performing pre-training, but the training heavily depends on how good and representative the unlabeled data is structured. The diversity and accessibility of the unlabeled dataset can affect how effective self-supervised learning can perform, if the unlabeled data used for training is not sufficiently diverse or representative of the target domain the TabTransformer variant's black-box nature i.e. Transformer may suffer with performance issues.

\section{Conclusion}
To conclude, our research has presented with a novel approach by creating various variant's of TabTransformer model namely, Binned-TT, Vanilla-MLP-TT, MLP-based-TT which has helped to increase the effective capturing of the underlying relationship between various features of the tabular dataset by constructing optimal inputs. And further we have employed a self-supervised learning approach in the form of a masking-based unsupervised setting for tabular data. 
\hfill\\\\
This approach has allowed us to train the TabTransformer without relying on labeled data, thereby eliminating the need for costly and time-consuming data annotation. Instead, our models learn the dependencies between features by leveraging the power of the Transformer's self-attention mechanism. This self-supervised learning approach has proven to be effective in capturing complex patterns and relationships within the tabular data as compared to supervised learning. During the experiments it was found that the SSL based MLP-TT was able to outperform all other variants and the base models for tabular data.
\hfill\\\\
A comparative analysis on the results have also been conducted to highlight the advantages of using self-supervised learning approach over the traditional supervised learning approach by comparing the results with the baseline model like MLP and TabTransformer in supervised settings. The comparison aimed to highlight the advantages of using self-supervised learning over traditional supervised learning. Our findings demonstrate that the self-supervised TabTransformer models consistently outperform the baseline models, showcasing the effectiveness of our approach in improving predictive performance. Overall, our research has contributed in advancing the original TabTransformer structure by creating optimal input construction for the TabTransformer to out perform the base models leveraging the self-supervised approach. 
\section{Future Scope}
There are several directions for future for our domain.  One direction is to explore the use of other self-supervised learning approaches for tabular data. For example, contrastive learning has been shown to be effective for natural language processing tasks. It would be interesting to investigate whether contrastive learning could also be effective for tabular data. Additionally, one can also incorporate other approach based on consistency based learning to examine it's reliability on tabular data. 
\hfill\\\\
Another direction for future research is to explore the use of other approaches to learning the data. For example, reinforcement learning has been shown to be effective for some tasks. It would be interesting to investigate whether reinforcement learning could also be effective for tabular data.
\hfill\\\\
While the majority of our research focuses on tabular data, it is important to consider whether our proposed methodology can also be applied to text and image data. Images have spatial relationships and frequently have rich semantic information that can benefit from self-supervised learning techniques. One might find new insights and confirm the effectiveness of our self-supervised TabTransformer structure in various domains by extending and adapting our approach to these different data types.



\appendix 


\chapter{Model Configurations and Datasets}
All model configurations and datasets have been attached in the following github repository:
\href{https://github.com/tirthvyas-tk-labs/Deep-Learning-with-Tabular-Data-A-Self-Supervised-Approach}{Deep Learning with Tabular Data: A Self-Supervised Approach}


\printbibliography[heading=bibintoc]

@article{c1,
  author={Jacob, Devlin and Ming-Wei, Chang and Kenton, Lee and Kristina, Toutanova},
  title={BERT: Pre-training of Deep Bidirectional Transformers for Language Understanding},
  journal={CoRR},
  volume={abs/1810.04805},
  year={2018},
  url={https://arxiv.org/abs/1810.04805},
  eprinttype={arXiv},
  eprint={1810.04805}
}

@article{c2,
  author={He, Kaiming and Zhang, Xiangyu and Ren, Shaoqing and Sun, Jian},
  booktitle={2016 IEEE Conference on Computer Vision and Pattern Recognition (CVPR)}, 
  title={Deep Residual Learning for Image Recognition}, 
  year={2016},
  volume={},
  number={},
  pages={770-778},
  doi={10.1109/CVPR.2016.90}
}

@article{c3,
  author={Deng, Jia and Dong, Wei and Socher, Richard and Li, Li-Jia and Kai Li and Li Fei-Fei},
  booktitle={2009 IEEE Conference on Computer Vision and Pattern Recognition}, 
  title={ImageNet: A large-scale hierarchical image database}, 
  year={2009},
  volume={},
  number={},
  pages={248-255},
  doi={10.1109/CVPR.2009.5206848}
}

@article{c4,
  author       = {Kaiming, He and
                  Xiangyu, Zhang and
                  Shaoqing, Ren and
                  Jian, Sun},
  title        = {Deep Residual Learning for Image Recognition},
  journal      = {CoRR},
  volume       = {abs/1512.03385},
  year         = {2015},
  url          = {http://arxiv.org/abs/1512.03385},
  eprinttype    = {arXiv},
  eprint       = {1512.03385},
  bibsource    = {dblp computer science bibliography, https://dblp.org}
}

@online{c5,
    title={OpenAI}, 
    url={https://openai.com/}, 
    journal={OpenAI}, 
    author={OpenAI}, 
    year={2019}, 
    month={Apr}
}

@online{c6,
    title={Google AI}, 
    url={https://ai.google/}, 
    journal={Google AI}, 
    author={Google}, 
    year={2019}
}

@article{c7,
  author       = {Tianqi, Chen and
                  Carlos, Guestrin},
  title        = {XGBoost: A Scalable Tree Boosting System},
  journal      = {CoRR},
  volume       = {abs/1603.02754},
  year         = {2016},
  url          = {http://arxiv.org/abs/1603.02754},
  eprinttype    = {arXiv},
  eprint       = {1603.02754},
  bibsource    = {dblp computer science bibliography, https://dblp.org}
}

@article{c8,
    title={Multimodal Auto{ML} on Structured Tables with Text Fields},
    author={Xingjian, Shi and Jonas, Mueller and Nick, Erickson and Mu, Li and Alex, Smola},
    booktitle={8th ICML Workshop on Automated Machine Learning (AutoML) },
    year={2021},
    url={https://openreview.net/forum?id=OHAIVOOl7Vl}
}

@article{c9,
      title={On Embeddings for Numerical Features in Tabular Deep Learning}, 
      author={Yury, Gorishniy and Ivan, Rubachev and Artem, Babenko},
      year={2022},
      eprint={2203.05556},
      archivePrefix={arXiv},
      primaryClass={cs.LG}
}

@article{c10,
  author       = {Doyen, Sahoo and
                  Quang, Pham and
                  Jing, Lu and
                  Steven, C. H. Hoi},
  title        = {Online Deep Learning: Learning Deep Neural Networks on the Fly},
  journal      = {CoRR},
  volume       = {abs/1711.03705},
  year         = {2017},
  url          = {http://arxiv.org/abs/1711.03705},
  eprinttype    = {arXiv},
  eprint       = {1711.03705},
  bibsource    = {dblp computer science bibliography, https://dblp.org}
}

@article{c11,
  title={Greedy function approximation: A gradient boosting machine.},
  author={Jerome H., Friedman},
  journal={Annals of Statistics},
  year={2001},
  volume={29},
  pages={1189-1232}
}

@article{c12,
author = {Breiman, Leo},
title = {Random Forests},
year = {2001},
issue_date = {October 1 2001},
publisher = {Kluwer Academic Publishers},
address = {USA},
volume = {45},
number = {1},
issn = {0885-6125},
url = {https://doi.org/10.1023/A:1010933404324},
doi = {10.1023/A:1010933404324},
journal = {Mach. Learn.},
month = {oct},
pages = {5–32},
numpages = {28}
}

@article{c13,
  title         = {A Neural Algorithm of Artistic Style},
  author        = {Leon, Gatys and Alexander, Ecker and Matthias, Bethge},
  year          = {2016},
  month         = {sep},
  journal       = {Journal of Vision},
  publisher     = {Association for Research in Vision and Ophthalmology ({ARVO})},
  volume        = {16},
  number        = {12},
  pages         = {326},
  doi           = {10.1167/16.12.326},
  url           = {https://doi.org/10.1167/16.12.326}
}

@article{c14,
  author       = {Wenzhe, Shi and
                  Jose, Caballero and
                  Ferenc, Husz{\'{a}}r and
                  Johannes, Totz and
                  Andrew P., Aitken and
                  Rob, Bishop and
                  Daniel, Rueckert and
                  Zehan, Wang},
  title        = {Real-Time Single Image and Video Super-Resolution Using an Efficient
                  Sub-Pixel Convolutional Neural Network},
  journal      = {CoRR},
  volume       = {abs/1609.05158},
  year         = {2016},
  url          = {http://arxiv.org/abs/1609.05158},
  eprinttype    = {arXiv},
  eprint       = {1609.05158},
  bibsource    = {dblp computer science bibliography, https://dblp.org}
}

@article{c15,
  author       = {Nicholas, Frosst and
                  Geoffrey E., Hinton},
  title        = {Distilling a Neural Network Into a Soft Decision Tree},
  journal      = {CoRR},
  volume       = {abs/1711.09784},
  year         = {2017},
  url          = {http://arxiv.org/abs/1711.09784},
  eprinttype    = {arXiv},
  eprint       = {1711.09784},
  bibsource    = {dblp computer science bibliography, https://dblp.org}
}

@article{c16,
author = {Covington, Paul and Adams, Jay and Sargin, Emre},
title = {Deep Neural Networks for YouTube Recommendations},
year = {2016},
isbn = {9781450340359},
publisher = {Association for Computing Machinery},
address = {New York, NY, USA},
url = {https://doi.org/10.1145/2959100.2959190},
doi = {10.1145/2959100.2959190},
pages = {191–198},
numpages = {8},
location = {Boston, Massachusetts, USA},
series = {RecSys '16}
}

@article{c17,
  author       = {Fuzhen, Zhuang and
                  Zhiyuan, Qi and
                  Keyu, Duan and
                  Dongbo, Xi and
                  Yongchun, Zhu and
                  Hengshu, Zhu and
                  Hui, Xiong and
                  Qing, He},
  title        = {A Comprehensive Survey on Transfer Learning},
  journal      = {CoRR},
  volume       = {abs/1911.02685},
  year         = {2019},
  url          = {http://arxiv.org/abs/1911.02685},
  eprinttype    = {arXiv},
  eprint       = {1911.02685},
  bibsource    = {dblp computer science bibliography, https://dblp.org}
}

@article{c18,
  author       = {Ronald, Richman and
                  Mario, V. W{\"{u}}thrich},
  title        = {LocalGLMnet: interpretable deep learning for tabular data},
  journal      = {CoRR},
  volume       = {abs/2107.11059},
  year         = {2021},
  url          = {https://arxiv.org/abs/2107.11059},
  eprinttype    = {arXiv},
  eprint       = {2107.11059},
  bibsource    = {dblp computer science bibliography, https://dblp.org}
}

@article{c19,
  title={LightGBM: A Highly Efficient Gradient Boosting Decision Tree},
  author={Guolin Ke and Qi Meng and Thomas Finley and Taifeng Wang and Wei Chen and Weidong Ma and Qiwei Ye and Tie-Yan Liu},
  booktitle={NIPS},
  year={2017}
}

@article{c20,
  author       = {Anna, Veronika Dorogush and
                  Andrey, Gulin and
                  Gleb, Gusev and
                  Nikita, Kazeev and
                  Liudmila, Ostroumova Prokhorenkova and
                  Aleksandr, Vorobev},
  title        = {Fighting biases with dynamic boosting},
  journal      = {CoRR},
  volume       = {abs/1706.09516},
  year         = {2017},
  url          = {http://arxiv.org/abs/1706.09516},
  eprinttype    = {arXiv},
  eprint       = {1706.09516},
  bibsource    = {dblp computer science bibliography, https://dblp.org}
}

@article{c21,
  title={Deep Neural Decision Forests},
  author={Peter Kontschieder and Madalina Fiterau and Antonio Criminisi and Samuel Rota Bul{\`o}},
  journal={2015 IEEE International Conference on Computer Vision (ICCV)},
  year={2015},
  pages={1467-1475}
}

@article{c22,
  author       = {Yongxin, Yang and
                  Irene, Garcia Morillo and
                  Timothy, M. Hospedales},
  title        = {Deep Neural Decision Trees},
  journal      = {CoRR},
  volume       = {abs/1806.06988},
  year         = {2018},
  url          = {http://arxiv.org/abs/1806.06988},
  eprinttype    = {arXiv},
  eprint       = {1806.06988},
  bibsource    = {dblp computer science bibliography, https://dblp.org}
}

@article{c23,
  title={Analysis of the AutoML Challenge Series 2015-2018},
  author={Isabelle, M Guyon and Lisheng, Sun-Hosoya and Marc, Boull{\'e} and Hugo, Jair Escalante and Sergio, Escalera and Zhengying, Liu and Damir, Jajetic and Bisakha, Ray and Mehreen, Saeed and Mich{\`e}le, Sebag and Alexander, R. Statnikov and Wei-Wei, Tu and Evelyne, Viegas},
  booktitle={Automated Machine Learning},
  year={2019}
}

@article{c24,
 author = {Beutel, Alex and Covington, Paul and Jain, Sagar and Xu, Can and Li, Jia and Gatto, Vince and Chi, Ed H},
  booktitle = {Proceedings of the Eleventh ACM International Conference on Web Search and Data Mining},
  organization = {ACM},
  pages = {46--54},
  publisher = {ACM},
  title = {Latent cross: Making use of context in recurrent recommender systems},
  url = {http://dblp.uni-trier.de/db/conf/wsdm/wsdm2018.html#BeutelCJXLGC18},
  year = 2018

}

@article{c25,
title	= {Are Neural Rankers still Outperformed by Gradient Boosted Decision Trees?},
author	= {Zhen, Qin and Le, Yan and Honglei, Zhuang and Yi, Tay and Rama, Kumar Pasumarthi and Xuanhui, Wang and Mike, Bendersky and Marc, Najork},
year	= {2021},
booktitle	= {International Conference on Learning Representations (ICLR)}
}

@article{c26,

  author       = {Ruoxi, Wang and
                  Bin, Fu and
                  Gang, Fu and
                  Mingliang, Wang},
  title        = {Deep {\&} Cross Network for Ad Click Predictions},
  journal      = {CoRR},
  volume       = {abs/1708.05123},
  year         = {2017},
  url          = {http://arxiv.org/abs/1708.05123},
  eprinttype    = {arXiv},
  eprint       = {1708.05123},
  bibsource    = {dblp computer science bibliography, https://dblp.org}

}

@article{c27,
  author       = {Xin, Huang and
                  Ashish, Khetan and
                  Milan, Cvitkovic and
                  Zohar, S. Karnin},
  title        = {TabTransformer: Tabular Data Modeling Using Contextual Embeddings},
  journal      = {CoRR},
  volume       = {abs/2012.06678},
  year         = {2020},
  url          = {https://arxiv.org/abs/2012.06678},
  eprinttype    = {arXiv},
  eprint       = {2012.06678},
  bibsource    = {dblp computer science bibliography, https://dblp.org}
}

@article{c28,
  title={Transfer Learning with Deep Tabular Models},
  author={Levin, Roman and Cherepanova, Valeriia and Schwarzschild, Avi and Bansal, Arpit and Bruss, C Bayan and Goldstein, Tom and Wilson, Andrew Gordon and Goldblum, Micah},
  journal={arXiv preprint arXiv:2206.15306},
  year={2022}
}

@article{c29,
  author       = {Yury, Gorishniy and
                  Ivan, Rubachev and
                  Valentin, Khrulkov and
                  Artem, Babenko},
  title        = {Revisiting Deep Learning Models for Tabular Data},
  journal      = {CoRR},
  volume       = {abs/2106.11959},
  year         = {2021},
  url          = {https://arxiv.org/abs/2106.11959},
  eprinttype    = {arXiv},
  eprint       = {2106.11959},
  bibsource    = {dblp computer science bibliography, https://dblp.org}
}

@article{c30,

  author       = {Ravid, Shwartz{-}Ziv and
                  Amitai, Armon},
  title        = {Tabular Data: Deep Learning is Not All You Need},
  journal      = {CoRR},
  volume       = {abs/2106.03253},
  year         = {2021},
  url          = {https://arxiv.org/abs/2106.03253},
  eprinttype    = {arXiv},
  eprint       = {2106.03253},
  bibsource    = {dblp computer science bibliography, https://dblp.org}

}

@article{c31,
  author       = {Sercan, {\"{O}}mer Arik and
                  Tomas, Pfister},
  title        = {TabNet: Attentive Interpretable Tabular Learning},
  journal      = {CoRR},
  volume       = {abs/1908.07442},
  year         = {2019},
  url          = {http://arxiv.org/abs/1908.07442},
  eprinttype    = {arXiv},
  eprint       = {1908.07442},
  bibsource    = {dblp computer science bibliography, https://dblp.org}

}

@article{c32,

  author       = {Sergei, Popov and
                  Stanislav, Morozov and
                  Artem, Babenko},
  title        = {Neural Oblivious Decision Ensembles for Deep Learning on Tabular Data},
  journal      = {CoRR},
  volume       = {abs/1909.06312},
  year         = {2019},
  url          = {http://arxiv.org/abs/1909.06312},
  eprinttype    = {arXiv},
  eprint       = {1909.06312},
  bibsource    = {dblp computer science bibliography, https://dblp.org}
  
}

@article{c33,

  title={Net-DNF: Effective Deep Modeling of Tabular Data},
  author={Liran, Katzir and Gal, Elidan and Ran, El-Yaniv},
  booktitle={International Conference on Learning Representations},
  year={2021}

}

@article{yoon,
author = {Yoon, Jinsung and Zhang, Yao and Jordon, James and van der Schaar, Mihaela},
title = {VIME: Extending the Success of Self- and Semi-Supervised Learning to Tabular Domain},
year = {2020},
isbn = {9781713829546},
publisher = {Curran Associates Inc.},
address = {Red Hook, NY, USA},
booktitle = {Proceedings of the 34th International Conference on Neural Information Processing Systems},
articleno = {926},
numpages = {11},
location = {Vancouver, BC, Canada},
series = {NIPS'20}
}

@article{ucar,
  author       = {Talip Ucar and
                  Ehsan Hajiramezanali and
                  Lindsay Edwards},
  title        = {SubTab: Subsetting Features of Tabular Data for Self-Supervised Representation
                  Learning},
  journal      = {CoRR},
  volume       = {abs/2110.04361},
  year         = {2021},
  url          = {https://arxiv.org/abs/2110.04361},
  eprinttype    = {arXiv},
  eprint       = {2110.04361},
  bibsource    = {dblp computer science bibliography, https://dblp.org}
}

@article{verma,

  author       = {Sahil, Verma and
                  John, P. Dickerson and
                  Keegan, Hines},
  title        = {Counterfactual Explanations for Machine Learning: Challenges Revisited},
  journal      = {CoRR},
  volume       = {abs/2106.07756},
  year         = {2021},
  url          = {https://arxiv.org/abs/2106.07756},
  eprinttype    = {arXiv},
  eprint       = {2106.07756},
  bibsource    = {dblp computer science bibliography, https://dblp.org}

}

@article{bahri,

  author       = {Dara Bahri and
                  Heinrich Jiang and
                  Yi Tay and
                  Donald Metzler},
  title        = {{SCARF:} Self-Supervised Contrastive Learning using Random Feature
                  Corruption},
  journal      = {CoRR},
  volume       = {abs/2106.15147},
  year         = {2021},
  url          = {https://arxiv.org/abs/2106.15147},
  eprinttype    = {arXiv},
  eprint       = {2106.15147},
  bibsource    = {dblp computer science bibliography, https://dblp.org}
}

@article{survey,

  author       = {Vadim, Borisov and
                  Tobias, Leemann and
                  Kathrin, Se{\ss}ler and
                  Johannes, Haug and
                  Martin, Pawelczyk and
                  Gjergji, Kasneci},
  title        = {Deep Neural Networks and Tabular Data: {A} Survey},
  journal      = {CoRR},
  volume       = {abs/2110.01889},
  year         = {2021},
  url          = {https://arxiv.org/abs/2110.01889},
  eprinttype    = {arXiv},
  eprint       = {2110.01889},
  bibsource    = {dblp computer science bibliography, https://dblp.org}
}

@article{atten, 
 title={Emerging Properties in Self-Supervised Vision Transformers}, 
 DOI={https://doi.org/10.1109/iccv48922.2021.00951}, journal={HAL (Le Centre pour la Communication Scientifique Directe)}, publisher={Le Centre pour la Communication Scientifique Directe}, author={Caron, Mathilde and Touvron, Hugo and Misra, Ishan and Hervé Jégou and Julien Mairal and Bojanowski, Piotr and Joulin, Armand}, year={2021}, month={Oct} }

@article{cnn_paper, title={CNN-based architecture for real-time object-oriented video coding applications}, volume={33}, DOI={https://doi.org/10.1002/cta.303}, number={1}, journal={International Journal of Circuit Theory and Applications}, author={Grassi, Giuseppe and Grieco, Luigi A.}, year={2005}, month={Jan}, pages={53–64} }

@article{tab_dl_problem, title={Explaining Deep Learning Models for Tabular Data Using Layer-Wise Relevance Propagation}, volume={12}, DOI={https://doi.org/10.3390/app12010136}, number={1}, journal={Applied Sciences}, author={Ullah, Ihsan and Rios, Andre and Gala, Vaibhav and Mckeever, Susan}, year={2021}, month={Dec}, pages={136} }

@article{ssl_bert, title={ALBERT: A Lite BERT for Self-supervised Learning of Language
  Representations}, DOI={https://doi.org/10.48550/arxiv.1909.11942}, journal={arXiv (Cornell University)}, publisher={Cornell University}, author={Lan, Zhenzhong and Chen, Mingda and Goodman, Sebastian and Gimpel, Kevin and Sharma, Piyush and Radu Soricut}, year={2020}, month={Apr} }

@article{ssl_vision, title={Emerging Properties in Self-Supervised Vision Transformers}, DOI={https://doi.org/10.1109/iccv48922.2021.00951}, journal={HAL (Le Centre pour la Communication Scientifique Directe)}, publisher={Le Centre pour la Communication Scientifique Directe}, author={Caron, Mathilde and Touvron, Hugo and Misra, Ishan and Hervé Jégou and Julien Mairal and Bojanowski, Piotr and Joulin, Armand}, year={2021}, month={Oct} }

@article{dl_paper, title={A Survey on Deep Learning its Architecture and Various Applications}, volume={1}, DOI={https://doi.org/10.21742/ajnnia.2017.1.2.02}, number={2}, journal={The Asia-Pacific Journal of Neural Networks and Its Applications}, author={Chandra, Praveen}, year={2017}, month={Oct}, pages={7–14} }

@article{tabt,
  author       = {Xin, Huang and
                  Ashish, Khetan and
                  Milan, Cvitkovic and
                  Zohar, S. Karnin},
  title        = {TabTransformer: Tabular Data Modeling Using Contextual Embeddings},
  journal      = {CoRR},
  volume       = {abs/2012.06678},
  year         = {2020},
  url          = {https://arxiv.org/abs/2012.06678},
  eprinttype    = {arXiv},
  eprint       = {2012.06678},
  bibsource    = {dblp computer science bibliography, https://dblp.org}
}

@inproceedings{cont_learn,
  title={Understanding contrastive learning requires incorporating inductive biases},
  author={Saunshi, Nikunj and Ash, Jordan and Goel, Surbhi and Misra, Dipendra and Zhang, Cyril and Arora, Sanjeev and Kakade, Sham and Krishnamurthy, Akshay},
  booktitle={International Conference on Machine Learning},
  pages={19250--19286},
  year={2022},
  organization={PMLR}
}

@article{mlm_learn,
  title={Learning Better Masking for Better Language Model Pre-training},
  author={Yang, Dongjie and Zhang, Zhuosheng and Zhao, Hai},
  journal={arXiv preprint arXiv:2208.10806},
  year={2022}
}

@article{autoenco,
  author       = {Michael, Tschannen and
                  Olivier, Bachem and
                  Mario, Lucic},
  title        = {Recent Advances in Autoencoder-Based Representation Learning},
  journal      = {CoRR},
  volume       = {abs/1812.05069},
  year         = {2018},
  url          = {http://arxiv.org/abs/1812.05069},
  eprinttype    = {arXiv},
  eprint       = {1812.05069},
  bibsource    = {dblp computer science bibliography, https://dblp.org}
}

@article{transformer,
  author       = {Ashish, Vaswani and
                  Noam, Shazeer and
                  Niki, Parmar and
                  Jakob, Uszkoreit and
                  Llion, Jones and
                  Aidan, N. Gomez and
                  Lukasz, Kaiser and
                  Illia, Polosukhin},
  title        = {Attention Is All You Need},
  journal      = {CoRR},
  volume       = {abs/1706.03762},
  year         = {2017},
  url          = {http://arxiv.org/abs/1706.03762},
  eprinttype    = {arXiv},
  eprint       = {1706.03762},
  bibsource    = {dblp computer science bibliography, https://dblp.org}
}

@misc{california_1997, title={California Housing Dataset}, url={https://www.dcc.fc.up.pt/~ltorgo/Regression/cal_housing.html}, journal={Fc.up.pt}, author={California}, year={1997} }

@misc{adult_data, url={https://archive.ics.uci.edu/ml/datasets/Adult},
author={Adult},
journal={Uci.edu}, year={2013} }

@misc{cancer_2011, title={UCI Machine Learning Repository: Breast Cancer Wisconsin (Diagnostic) Data Set}, url={https://archive.ics.uci.edu/ml/datasets/Breast+Cancer+Wisconsin+(Diagnostic)}, journal={Uci.edu}, author={Wisconsin Cancer Data}, year={2011} }


\end{document}